\documentclass[journal]{IEEEtran}
\usepackage{times}
\usepackage{epsfig}
\usepackage{graphicx}
\usepackage{amsmath}
\usepackage{amssymb}
\usepackage{mathtools}
\usepackage{algorithm}
\usepackage{algorithmic}
\usepackage{graphics}
\usepackage{subfigure}
\usepackage{multirow}
\usepackage{booktabs}
\usepackage{url}
\usepackage{amsthm}
\usepackage{wrapfig}
\usepackage{xcolor}
\usepackage{utfsym}
\usepackage[normalem]{ulem}
\usepackage{makecell}
\usepackage{ragged2e}

\usepackage{tablefootnote}

\usepackage{amssymb}

\usepackage{xspace}

\usepackage{cite}

\DeclareMathOperator*{\argmax}{argmax}
\DeclareMathOperator*{\argmin}{argmin}

\begin{document}
\title{Filter Pruning for Efficient CNNs via Knowledge-driven Differential Filter Sampler}

\author{Shaohui~Lin,
    Wenxuan~Huang,
    Jiao~Xie$^{*}$,
    Baochang~Zhang,
    Yunhang~Shen,
    Zhou~Yu,
    Jungong~Han,
    and David~Doermann,~\IEEEmembership{Fellow,~IEEE}
    \thanks{S. Lin is with the School of Computer Science and Technology, East China Normal University, Shanghai, China, and the Key Laboratory of Advanced Theory and Application in Statistics and Data Science, Ministry of Education, China. (shlin@cs.ecnu.edu.cn)}
    \thanks{W. Huang is with the School of Computer Science and Technology, East China Normal University, Shanghai, China. (osilly0616@gmail.com)}
    \thanks{J. Xie is with the Department of Automation, School of Aerospace Engineering, Xiamen University, Xiamen, China. (jiaoxie1990@126.com)}
    \thanks{B. Zhang is with the Institute of Artificial Intelligence, Beihang University, Beijing, China, and Zhongguancun Laboratory, Beijing, China. (bczhang@buaa.edu.cn)}
    \thanks{Y. Shen is with the Youtu Lab, Tencent, Shanghai, China. (shenyunhang01@gmail.com)}
    \thanks{Y. Zhou is with the School of Statistics, East China Normal University, Shanghai, China, and the Key Laboratory of Advanced Theory and Application in Statistics and Data Science, Ministry of Education, China. (zyu@stat.ecnu.edu.cn)}
    \thanks{J. Han is with the Department of Computer Science, University of Sheffield, UK. (jungonghan77@gmail.com)}
    \thanks{D. Doermann is with the University at Buffalo, Buffalo, NY, USA. (doermann@buffalo.edu)}
    \thanks{*Corresponding author: Jiao Xie}
}

%
%
\markboth{Journal of \LaTeX\ Class Files,~Vol.~14, No.~8, August~2021}%
{Shell \MakeLowercase{\textit{et al.}}: Bare Demo of IEEEtran.cls for IEEE Journals}

\maketitle

\begin{abstract}
    Filter pruning simultaneously accelerates the computation and reduces the memory overhead of CNNs, which can be effectively applied to edge devices and cloud services.
    In this paper, we propose a novel Knowledge-driven Differential Filter Sampler~(KDFS) with Masked Filter Modeling~(MFM) framework for filter pruning, which globally prunes the redundant filters based on the prior knowledge of a pre-trained model in a differential and non-alternative optimization.
    Specifically, we design a differential sampler with learnable sampling parameters to build a binary mask vector for each layer, determining whether the corresponding filters are redundant.
    To learn the mask, we introduce masked filter modeling to construct PCA-like knowledge by aligning the intermediate features from the pre-trained teacher model and the outputs of the student decoder taking sampling features as the input.
    The mask and sampler are directly optimized by the Gumbel-Softmax Straight-Through Gradient Estimator in an end-to-end manner in combination with global pruning constraint, MFM reconstruction error, and dark knowledge.
    Extensive experiments demonstrate the proposed KDFS's effectiveness in compressing the base models on various datasets.
    For instance, the pruned ResNet-50 on ImageNet achieves $55.36\%$ computation reduction, and $42.86\%$ parameter reduction, while only dropping $0.35\%$ Top-1 accuracy, significantly outperforming the state-of-the-art methods.
    The code is available at \url{https://github.com/Osilly/KDFS}.
\end{abstract}


\begin{figure*}[t]
    \centering
    \includegraphics[scale = 0.38]{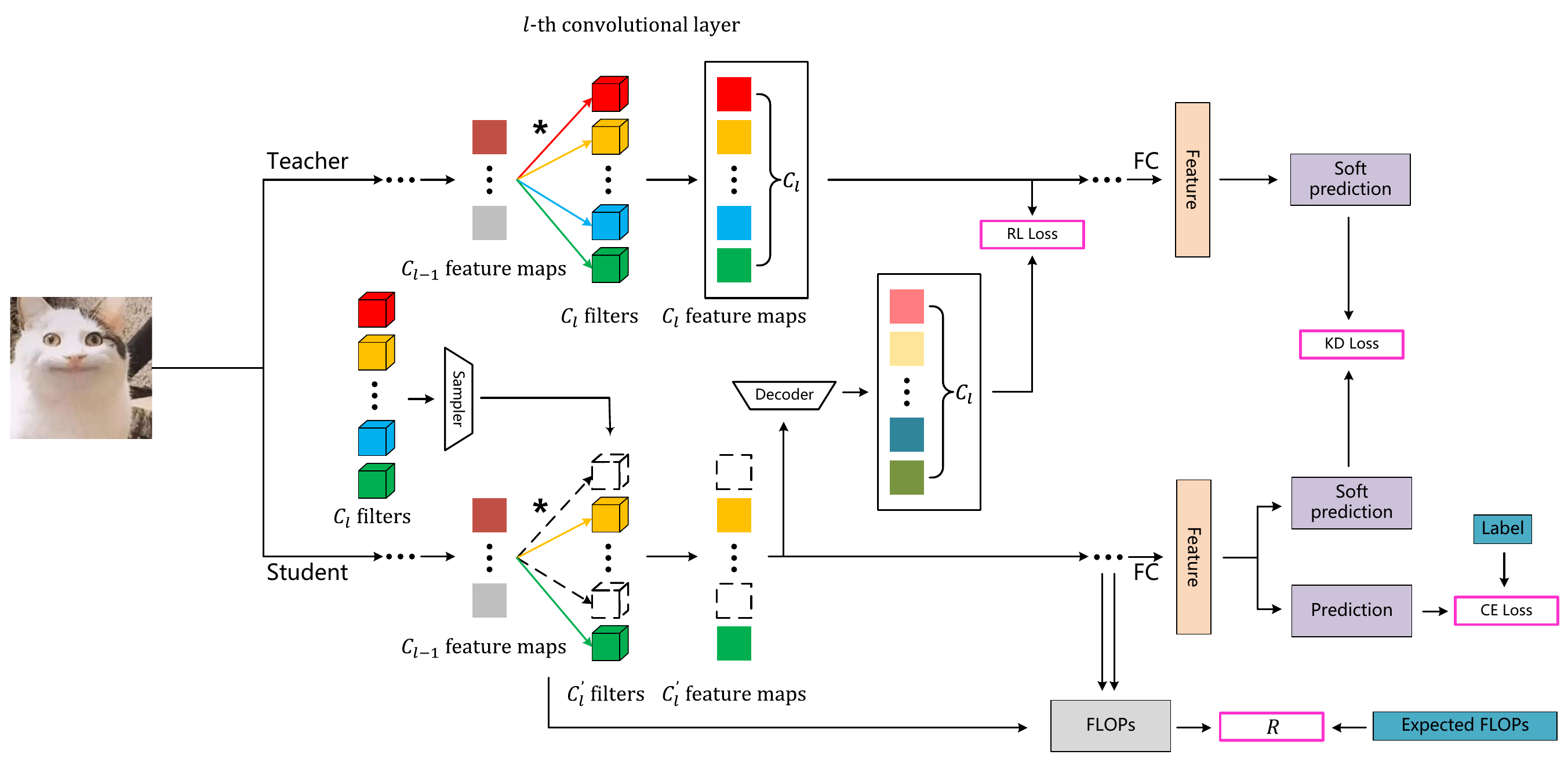}
    \caption{The illustration of KDFS. Its key components lie in differential filter sampler, masked filter modeling between teacher output and student decoder, and FLOPs regularization term $\mathcal{R}$.
    The sampler with additional sampling parameters is proposed to generate binary masks to automatically select the filters, where dotted cubes denote the mask values are $0$.
    To better guide the sampling, masked filter modeling exploits the prior knowledge from a pre-trained model (\emph{a.k.a.} teacher) to construct PCA-like knowledge (\emph{i.e.}, RL loss), which aligns the teacher intermediate features and the outputs of student decoder taking sampling features as the input.
    We leverage FLOPs regularization $\mathcal{R}$ into RL loss, CE loss, and KD loss and optimize the sampler and weights directly in an end-to-end manner.
    }
    \label{fig1}
\end{figure*}

\section{Introduction}
\label{intro}
\IEEEPARstart{C}{onvolutional} Neural Networks~(CNNs) have achieved remarkable performance on computer vision, such as image classification~\cite{krizhevsky2012imagenet,simonyan2015very,szegedy2015going,he2016deep,huang2017densely}, object detection~\cite{girshick2014rich,girshick2015fast,ren2015faster,he2017mask}, and image segmentation~\cite{long2015fully,chen2017deeplab,chen2017rethinking}, \emph{etc}. Performance boosting of CNNs is often accompanied by significant computation and memory consumption, which restricts their applications on resource-limited edge devices and cloud services. For edge computing like autonomous driving, huge computation overhead undermines user experience and causes safety concerns. Even for cloud service, large-capacity models could be compressed to improve computation throughput, allowing more clients and users to reduce computing costs and waiting time with a high-quality experience. Network pruning is not only effective and efficient in reducing model complexity but also compatible with different CNN compression methods, such as low-rank decomposition \cite{lin2018holistic,li2021towards,zhang2015accelerating,denton2014exploiting}, knowledge distillation~\cite{hinton2015distilling,romero2014fitnets} and quantization~\cite{rastegari2016xnor,jacob2018quantization,zhao2022towards}. As such, network pruning has received a great deal of research focus both in industry and academia.

Network pruning removes either the fine-grained filter weights (\emph{a.k.a.} weight pruning)~\cite{han2015learning,han2016deep,frankle2018lottery} or the entire filters (\emph{a.k.a.} filter pruning)~\cite{lin2019towards,li2017pruning,liu2017learning,lin2020hrank,he2019filter} to obtain sparse models for compression and acceleration.
%
However, weight pruning may reduce the practical inference speed due to irregular memory access, such that it has to rely on specialized software~\cite{park2017faster} or hardware~\cite{han2016eie} for fast inference.
In contrast, filter or channel pruning is more friendly to implement fast inference without the software and hardware constraints. This paper focuses on filter pruning to provide a versatile solution to achieve a better trade-off between computation and accuracy for different devices.

According to the pruning way, filter pruning can be further categorized into \emph{hard filter pruning} and \emph{soft filter pruning}.
Hard filter pruning always \emph{fixes} the pruned filters during the whole training process.
For instance, He~\emph{et al.}~\cite{he2017channel} and Luo~\emph{et al.}~\cite{luo2017thinet} pruned filters and the corresponding feature maps by considering statistics computed from the next layer in a greedy layer-wise manner.
Magnitude-based pruning methods proposed the L1-norm of filter~\cite{li2017pruning}, the sparsity of feature maps~\cite{hu2016network}, or average rank of feature maps~\cite{lin2020hrank} to determine the filter saliency. Then, they iteratively prune the lowest salient filters and retrain the pruned network layer-by-layer.
However, the mistakenly pruned filters cannot be rolled back for training, which leads to the low discriminative ability of the final pruned models.
Moreover, these methods require manually predefining a pruning rate for each layer, which neglects the layer pruning sensitivity~\cite{lin2019toward} and decreases the pruning flexibility.
Soft filter pruning aims to \emph{dynamically} prune redundant filters during training, which enables the networks to determine the filter importance adaptively via group sparsity-based weight regularization~\cite{lin2019toward,zhou2016less,wen2016learning}, sparsity of batch-normalization~(BN) scalars~\cite{liu2017learning,ye2018rethinking,you2019gate}, or binary mask~\cite{huang2018data,yu2018nisp,lin2019towards,gao2022disentangled,lin2018accelerating,gao2020discrete}.
Group sparsity-based regularization usually imposes $L_{2,1}$-norm~\cite{zhou2016less,wen2016learning} and $L_{2,0}$-norm~\cite{lin2019toward} constraints on weights, making filters sparse automatically during training.
To further determine the redundant filters, sparse regularization on the scaling parameters in BN layer~\cite{liu2017learning,you2019gate} is introduced to implement sparse scalars.
However, they require a trial-and-error procedure to obtain the predetermined pruning rate.
As a solution, several works~\cite{huang2018data,lin2019towards,lin2018accelerating,gao2020discrete,guo2021gdp} learned binary masks to determine the redundant filters by an alternative optimization between filters and masks or the specific architecture~\cite{gao2022disentangled}.
However, the alternative optimization significantly decreases the training efficiency. It fails to exploit the knowledge from pre-trained models to guide the filter pruning, which may lead to sub-optimal compression results.

To address the above problems, we propose a novel soft filter pruning scheme via a Knowledge-driven Differential Filter Sampler~(termed \emph{KDFS}) with Masked Filter Modeling~(MFM), which globally prunes the redundant filters learned from the knowledge of pre-trained model in a differential and non-alternative optimization process.
Fig.~\ref{fig1} depicts the workflow of the proposed approach.
In particular, we would like to propose a differential sampler to build a binary mask to determine the redundant filters during training automatically.
Inspired by MAE~\cite{he2022masked}, we then introduce masked filter modeling to construct PCA-like knowledge by aligning the outputs between the intermediate features of the pre-trained teacher and the decoder added to the student, which guides the filter sampling based on the Straight-Through Gradient Estimator.
Instead of sparsity constraint on the binary mask, we further design a FLOPs regularization term, which is leveraged into knowledge distillation to learn the student with end-to-end optimization.

The main contributions can be summarized as follows:
\begin{itemize}
    \item We propose a knowledge-driven differential filter sampler~(KDFS) to remove redundant filters for efficient CNN inference. From a global pruning perspective, the unimportant filters in each layer are automatically determined by a binary mask and additional Gumbel-Softmax sampling parameters.
    \item We exploit the intermediate knowledge from a pre-trained model and propose a masked filter modeling~(MFM) to construct PCA-like knowledge for better guiding the learning of the student network, which is also leveraged into the traditional knowledge distillation for performance improvement. Instead of the sparsity constraint on the binary mask, we introduce the FLOPs regularization term to force the pruned network to achieve the pre-defined global pruning rate without trial and error.
    \item Extensive experiments demonstrate the superior performance of the proposed KDFS scheme. On ImageNet ILSVRC 2012, the pruned ResNet-50 achieves $55.36\%$ computation reduction, and $42.86\%$ parameter reduction, while its Top-1 accuracy only drops $0.35\%$ outperforming the state-of-the-art pruning methods.
\end{itemize}

The rest of this paper is organized as follows: In Section~\ref{rw}, related works are introduced and discussed. The proposed KDFS method is described in Section~\ref{method}, including the preliminaries in Subsection~\ref{ssub_pf}, the method overview in Subsection~\ref{overview}, its critical components in Subsection~\ref{components} and optimization in Subsection~\ref{opt}. Elaborate experiments and analysis are conducted in Section~\ref{exp}. Finally, the summary of the proposed KDFS is presented in Section~\ref{con}.



\section{Related Work}
\label{rw}
\subsection{Network Pruning}
Network pruning aims to remove network connections in a non-structured or structured manner. Early works in non-structured pruning (or weight pruning) proposed saliency measurements to remove redundant weights determined by the second-order derivative matrix of the loss function \emph{w.r.t.} the weights~\cite{hassibi1993second,lecun1989optimal} or the absolute values of weights~\cite{han2015learning,han2016deep,frankle2018lottery}. To avoid incorrect weight pruning, Guo~\emph{et al.}~\cite{guo2016dynamic} proposed a connection splicing to reduce the accuracy loss of the pruned network. In contrast, structured pruning can reduce the network size for fast inference without specialized packages. Channel or filter pruning is one of the most widely-used structured ways for CNN acceleration and compression, which can be further categorized into hard filter pruning and soft filter pruning. Hard filter pruning aims to prune filters during training in a fixed and static manner, which designs an important score of filters to iteratively prune the ``least important'' filters and retrain the pruned network layer-by-layer. L1-norm of filters~\cite{li2017pruning}, Geometric Median of filters~\cite{he2019filter} and the statistics of feature maps~\cite{he2017channel,luo2017thinet,hu2016network,lin2020hrank} can be used to determine the filter importance. Recently, importance measurements based on the first-order Taylor expansion~\cite{molchanov2017pruning,molchanov2019importance} and second-order derivatives~\cite{dong2017learning,peng2019collaborative,zheng2022savit} have been proposed to approximate the loss of pruning.
However, these hard pruning methods are inadaptive in that mistakenly pruning the salient filters is irretrievable to decrease the discriminative ability of the final pruned models.

In line with our work, soft filter pruning prunes redundant filters dynamically and adaptively, which allows the pruned networks to restore the mistakenly pruned filters.
Works in~\cite{lin2019toward,zhou2016less,wen2016learning} proposed group sparsity regularization of weights to penalize unimportant parameters and prune redundant filters during training.
Several works~\cite{liu2017learning,you2019gate} found that scaling factors in batch normalization~(BN) are associated with filters and imposed $L1-$regularization on scaling factors in BN to identify and prune unimportant channels. However, these regularization methods require a certain threshold to achieve exact zero values and train the network in a trial-and-error manner to achieve the pruned network with a pre-defined compression rate. Recently, binary masks have been introduced to determine filter importance, which is learned by either alternating optimization~\cite{lin2019towards,huang2018data,lin2018accelerating,gao2020discrete,guo2021gdp} or the specific architecture~\cite{gao2022disentangled}. For example, Lin~\emph{et al.}~\cite{lin2018accelerating} proposed a global and dynamic pruning method to reduce redundant filters by greedy alternative updating between weights and binary masks. Discrete gate~\cite{gao2020discrete} searched a sub-network with specific channels, which is optimized by the straight-through estimator~(STE)~\cite{bengio2013estimating}.
Unlike these methods, we design a differential filter sampler to estimate the binary masks and introduce masked filter modeling to construct PCA-like knowledge for distilling the pre-trained model, which better guides the pruning direction and simultaneously optimizes weights and binary masks in an end-to-end manner.

Note that SE-based gates~\cite{hu2018squeeze,veit2018convolutional,chen2019you,gao2018dynamic} have been used to dynamically prune the network for adaptive inference, where Gumbel-Max trick is used for the propagation of gradients during training. However, these methods require more parameters and computations to obtain the gates (\emph{e.g.}, in FBS~\cite{veit2018convolutional}) for filter selection during both training and inference, compared to our sampler. Specifically, our sampler is safely removed without computation overhead for inference.

\subsection{Neural Architecture Search}
The previous work~\cite{liu2018rethinking} revealed that the pruned network architecture is essential to determine the pruning performance. Therefore, automatic search of neural architecture has been used in filter pruning by reinforcement learning~\cite{he2018amc}, employed bee~\cite{lin2021channel} or Q-value~\cite{lin2017runtime}, which significantly improves the performance of neural networks. However, the search space of these methods is large, which requires significant computational overhead to search and select the best model from hundreds of model candidates. In contrast, our method learns a compact neural architecture with weights from a single training, which is more efficient.

\subsection{Knowledge Distillation}
Our PCA-like knowledge is also related to knowledge distillation to a certain extent. Hinton~\emph{et al.}~\cite{hinton2015distilling} proposed dark knowledge by forcing the softened output logits of the student model to imitate that of the teacher model during training. Recently, many methods further exploited the teacher-student alignment of intermediate feature maps~\cite{romero2014fitnets}, instance relational graphs~\cite{liu2019knowledge}, similarity attention maps~\cite{ji2021show,zagoruyko2016paying} and generative adversarial predictions~\cite{micaelli2019zero}. Zhuang~\emph{et al.}~\cite{zhuang2018discrimination} combined the discrimination-aware losses and reconstruction error in the intermediate layers to select the most discriminative channels using a greedy algorithm layer-by-layer. Unlike the greedy process of filter selection, our method automatically determines the redundant filters at once from a global perspective, which leverages PCA-like knowledge from masked filters into dark knowledge. In addition, different from MAE~\cite{he2022masked} through reconstructing the masked image patches to learn feature representation, our PCA-like knowledge is extracted by reconstructing the features of the student to approach those of the pre-trained teacher. 

\begin{figure*}[t]
    \centering
    \includegraphics[scale = 0.5]{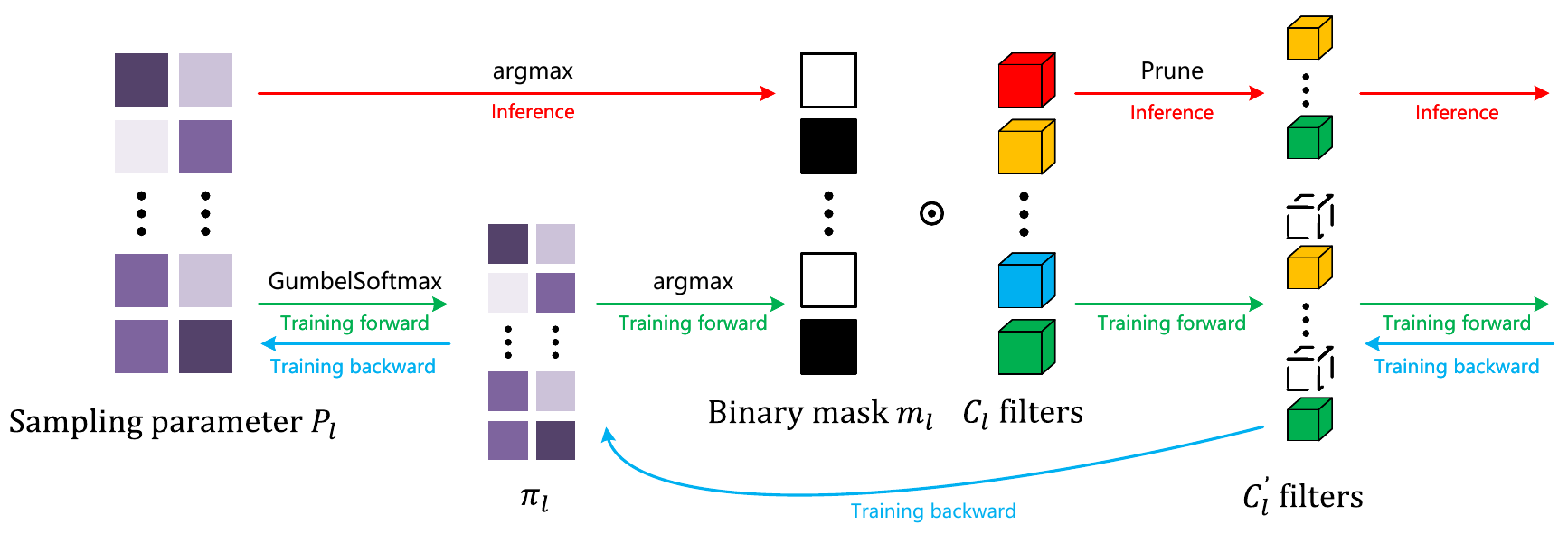}
    \caption{Pipeline of Gumbel-Softmax Straight-Through Gradient Estimator. Red, green, and blue arrows present the inference, training forward, and training backward processes, respectively.}
    \label{fig2}
\end{figure*}

\section{Methodology}
\label{method}

\subsection{Notations and Preliminaries}
\label{ssub_pf}

Consider a convolutional neural network, which maps the input image to a specific output class prediction.
We denote a set of feature maps in the $l$-th convolutional layer by $\mathcal{Z}_l\in\mathbb{R}^{H_l\times W_l\times C_l}$, where $H_l \times W_l$ and $C_l$ are spatial size and the number of individual maps (or channels).
The feature maps can either be the input of the network $\mathcal{Z}_0$, or the output feature maps $\mathcal{Z}_l$ with $l\in [1,2,\cdots, L]$.
We further denote individual feature map as ${\mathbf{Z}_{l}^{i}}\in{\mathbb{R}^{H_l\times W_l}}$ with $i\in [1,2,\cdots, C_l]$.
The convolutional operator ($*$) is applied to a set of input feature maps with filters parameterized by ${\mathcal{W}_{l}^{i}}\in{{\mathbb{R}}^{{d}\times{d}\times{C_{l-1}}}}$ to compute $\mathbf{Z}_{l}^{i}$, \emph{i.e.},
\begin{equation}
    \label{eq4}
    \mathbf{Z}_l^{i} = f(\mathcal{Z}_{l-1}*\mathcal{W}_l^{i}),
\end{equation}
where $f(\cdot)$ is a non-linear activation function, \emph{e.g.}, rectifier linear unit~(ReLU). For simplicity, we discuss the problem without the bias term and the batch normalization~(BN) layers.
By feed-forwarding all layers, we obtain the final prediction $\hat{y}$ as:
\begin{equation}
    \label{eq_y}
    \hat{y} = \text{softmax}\big(FC(f_p(\mathcal{Z}_{L}))\big).
\end{equation}
where $\text{softmax}$, $FC$ and $f_p$ are the softmax operation, fully-connected layer, and pooling layer, respectively. By putting all filter parameters together, we get the filter parameter set $\mathcal{W}=\{\{\mathcal{W}_l^i\}_{i=1}^{C_l} \}_{l=1}^{L}$, where $C=\sum_{l=1}^{L}C_l$ is the total number of filters in the network. In this paper, we need to construct the knowledge from the pre-trained model, thus regarding the pre-trained model as the teacher, whose pre-trained weight set and feature maps at $l$-th layer are fixed and denoted as $\mathcal{W}^{(t)}$ and $\mathcal{Z}_{l}^{(t)}$, respectively.

For the $l$-th convolutional layer, we consider to keep a number of $C_{l}^{'}$ filters or output feature maps (\emph{i.e.}, remove $C_{l}-C_{l}^{'}$ filters).
Therefore, we need to learn the subset $S_l \subseteq \{1,2,\cdots,C_{l}\}$ of filters with a specified size $|S_l|=C_{l}\times (1-r_l)=C_{l}^{'}$ ($r_l$ is a compression rate at the $l$-th layer), which is used to remove the corresponding redundant filters.
To implement the above filter selection, we introduce a binary mask $\mathbf{m}_l$ of size $C_l$ to determine the filters at the $l$-th layer.
Such that, the $i$-th output of $l$-th convolution is expressed as:
\begin{equation}
    \label{eq5}
    \mathbf{Z}_{l}^{i} = f(\mathbf{Z}_{l-1}*(\mathcal{W}_l^{i}\cdot\mathbf{m}_{l}^{i})), \quad s.t. \,\  \|\mathbf{m}_{l}\|_0 \leq |S_l|,
\end{equation}
\noindent where $\mathbf{m}_{l}^{i}$ is a binary value and is evaluated to $1$ if the $i$-th filter is determined to be preserved and $0$ otherwise.
Given a training set $\mathcal{D} = {(\mathbf{x}_i, y_i)}_{i=1}^{N}$, the problem of filter pruning is generally formulated as the following optimization problem:
\begin{equation}
    \label{eq_gen}
    \begin{split}   & \argmin_{\mathcal{W},\mathbf{m}}\frac{1}{N}\sum_{i=1}^{N}\mathcal{L}_{ce}(y_i,\phi(\mathbf{x_i};\mathcal{W},\mathbf{m})), \\
        & s.t. \quad \|\mathbf{m}\|_0\leq\sum_{l=1}^{L}|S_l|,
    \end{split}
\end{equation}
where $\mathcal{L}_{ce}(\cdot,\cdot)$ is the cross-entropy loss and  $\hat{y}_i = \phi(\mathbf{x_i},\mathcal{W},\mathbf{m})$ is the final network prediction. We denote the network with the binary mask as the student for a better description of our method.

To solve Eq.~\ref{eq_gen}, the previous works~\cite{li2017pruning,luo2017thinet,he2017channel} rely on the score function of filter importance and progressively remove the redundant filter layer-by-layer with a pre-defined compression rate for each layer, which decreases the pruning flexibility.
%
Alternatively, iterative optimization between filters and masks~\cite{gao2022disentangled,lin2018accelerating,gao2020discrete}, and the soft mask with the L1-norm constraint~\cite{huang2018data,lin2019towards} are introduced to simplify the non-convex optimization, which is mainly due to the non-differential binary mask.

We introduce a simple sampling parameter matrix for sub-network selection during training instead of using an additional importance score network in~\cite{gao2022disentangled}. In particular,
%
we randomly initialize a sampling parameter matrix $\mathbf{P}_l\in \mathbb{R}^{C_l\times 2}$ with the Kaiming Initialization~\cite{he2015delving} for the $l$-th convolutional layer, and the binary mask $\mathbf{m}_l$ is determined as:
\begin{equation}
    \label{eq6}
    \mathbf{m}_l^i=\left \{ \begin{matrix}
        1, & \argmax_{j}{\mathbf{P}_{l}^{i,j}} = 2, \\
        0, & \text{otherwise},                      \\
    \end{matrix}
    \right.
\end{equation}
\noindent where $i\in [1,2,\cdots, {C_l}]$ and $j\in [1,2]$.
Eq.~\ref{eq6} indicates that we sample values of the sampling parameter matrix $\mathbf{P}_l$ along the second dimension. However, directly sampling $\mathbf{m}_l$ from $\mathbf{P}_l$ is non-differentiable and impedes end-to-end training.
In addition, we noted that the pre-trained model is regarded as a teacher to improve the effectiveness of filter sampling using its prior knowledge.
Therefore, we propose a knowledge-driven differential filter sampler~(KDFS) to optimize the filter and binary mask simultaneously and replace the L$0$-norm constraint on the binary mask with FLOPs constraint with a global pruning rate $r$. In the following Subsection, we will describe the detailed implementation of KDFS.

\subsection{Method Overview}
\label{overview}
As discussed above in Subsection~\ref{ssub_pf}, we first relax the discrete binary mask $\mathbf{m}_l$ into a continuous random variable matrix computed by the Straight-Through Gumbel Softmax function~\cite{jang2016categorical} on $\mathbf{P}_l$ during training. During inference, the discrete binary mask $\mathbf{m}_l$ is used to remove redundant filters, and the network with binary mask and sampling parameters is regarded as a student.
Furthermore, we leverage the knowledge of the pre-trained teacher model into the student to construct PCA-like knowledge, which guides the filter sampling to improve the discriminative ability of the student. Finally, we set a simple yet effective FLOPs constraint with a global pruning rate and introduce the knowledge distillation to optimize the student network. The entire pipeline is presented in Fig.~\ref{fig1}, and the total objective function $\mathcal{J}$ is formulated as:
\begin{equation}
    \footnotesize
    \label{eq_all}
    \begin{split}
        &\argmin_{\mathcal{W},\theta,\mathbf{P}}\,  \frac{1}{N}\sum_{i=1}^{N}\mathcal{L}_{CE}\big(y_i,\phi(\mathbf{x}_i;\mathcal{W},\mathbf{m}(\mathbf{P}))\big) + \lambda_{1}\sum_{i=1}^{N}\mathcal{L}_{KD}\big(\mathbf{a}_i^{(t)}, \mathbf{a}_i^{(s)}\big) \\
        & + \lambda_{2}\sum_{l=1}^{L}{\mathcal{L}_{RL}}\big(\mathcal{Z}_l^{(t)}, D_{\theta_l}(\mathcal{Z}_l^{(s)}; \theta_l)\big) + \lambda_{3}{\mathcal{R}}\big(\psi(\mathcal{W},\mathbf{m}(\mathbf{P})), \psi(\mathcal{W}^{(t)})\big),
    \end{split}
\end{equation}
where $\mathcal{L}_{CE}, \mathcal{L}_{KD}, \mathcal{L}_{RL}$ and $\mathcal{R}$ are cross-entropy loss, knowledge distillation loss, reconstruction loss of masked filter modeling, and a FLOPs regularization term, respectively. $\mathbf{a}^{(t)}$ and $\mathbf{a}^{(s)}$ present the soft outputs of the teacher and student with a softened factor $T$~\cite{hinton2015distilling}, respectively.
$D_{\theta_l}(\mathcal{Z}_l^{(s)}, \theta_l)$ is the reconstructed features from a decoder $D$ with parameter $\theta_l$ at the $l$-th layer, which has the same size as the teacher's feature maps $\mathcal{Z}_l^{(t)}$.
$\psi$ defines the computation or parameter complexity, such as FLOPs. $\lambda_{1}$, $\lambda_{2}$ and $\lambda_{3}$ are the hyperparameters.
The following will describe how to minimize Eq.~\ref{eq_all} and the key components, such as the differential filter sampler, masked filter modeling, and FLOPs regularization term.

\subsection{Key Components of KDFS}
\label{components}
\textbf{Differential filter sampler~(DFS).}
We introduce the Straight-Through Gradient Estimator technique~\cite{bengio2013estimating} to approximate gradients \emph{w.r.t} binary mask.
In particular, we discretize the Gumbel Softmax sampling parameters by argmax operation in the forward propagation while keeping them continuous for back-propagation, ensuring that the training and inference dynamics remain consistent.
As shown in Fig.~\ref{fig2}, the sampling step during training and inference can be formulated as follows:

In the forward propagation, we aim to obtain categorical distribution to approximate the binary mask, such that the Gumbel-Softmax function is applied to the sampling parameter matrix $\mathbf{P}_{l}$:
\begin{equation}
    \label{eq7}
    \begin{aligned}
        \mathbf{\pi}_{l}^{(i,k)} & = \text{GumbelSoftmax}(\mathbf{P}_{l}^{(i,k)}, \mathbf{G}_{l}^{(i,k)})                                                                                                        \\
                                 & =\frac{\text{exp}\big[(\mathbf{P}_{l}^{(i,k)} + \mathbf{G}_{l}^{(i,k)})/\tau\big]}{\sum_{j=1}^{2}\text{exp}\big[(\mathbf{P}_{l}^{(i,j)} + \mathbf{G}_{l}^{(i,j)})/\tau\big]},
    \end{aligned}
\end{equation}
\noindent where $\mathbf{G}_{l}^{(i,k)}\sim \text{Gumbel}(0,1)$ is a random noise following Gumbel distribution.
If the temperature $\tau\rightarrow 0$, the discrete random variable $\mathbf{\pi}_{l}^{(i,k)}$ smoothly approaches the discrete distribution.
As $\tau$ becomes larger, $\mathbf{\pi}_{l}^{(i,k)}$ becomes a continuous random variable with more approaching to the uniform distribution. Then, we apply the argmax operation along the second dimension of $\mathbf{\pi}_{l}$ to obtain the binary mask:
\begin{equation}
    \label{eq8}
    \mathbf{m}_l^i=\left \{ \begin{matrix}
        1, & \argmax_j{\pi_{l}^{i,j}} = 2, \\
        0, & \text{otherwise}.             \\
    \end{matrix}
    \right.
\end{equation}
Therefore, Eq.~\ref{eq5} without restricting L0-norm is used to compute the sampling outputs layer-by-layer.

In practice, we employ an annealing schedule for the temperature $\tau$ in Eq.~\ref{eq7}.
Let $\tau_0$ and $\tau_E$ denote the beginning and final temperatures with the total number of epochs $E$, respectively.
For the decay, we set the current temperature at epoch number $e$ as $\tau(e)=(1-\frac{e}{E})\tau_0+\frac{e}{E}\tau_E$.
For exponential decay, the current temperature at epoch number $e$ is set to $\tau(e)=\tau_0(\tau_E/\tau_0)^{e/E}$.
By gradually decreasing $\tau$, $\mathbf{m}_l$ is closer to a true one-hot vector, and the gap between $\pi$ and $\mathbf{m}$ is reducing, especially in the late training.

For backward propagation, we formulate the gradients of $\pi_l$ as:
\begin{equation}
    \label{eq9}
    \frac{\partial{\cal{L}}}{\partial{\mathbf{\pi}_l}} = \frac{\partial{\cal{L}}}{\partial{\mathbf{m}_l}}.
\end{equation}
It indicates that we propagate the gradient flow from $\mathbf{m}_l$ to $\mathbf{\pi}_l$ using the Straight-Through Gradient Estimator.

For inference, we use the argmax operation on sampling parameter $\mathbf{P}$ to directly obtain the binary mask, which is similar to forward propagation without Gumbel Softmax.

\textbf{Masked filter modeling~(MFM).}
Directly adding the differential filter sampler to the student may lead to sub-optimal solutions due to the need for more guidance of local knowledge.
Alternatively, we can align the teacher's feature maps and the student sampler outputs to introduce the local knowledge for optimizing the student.
However, it results in a significant performance drop, as the output feature representation between the student and the teacher is mismatched when the filters are sampled.
To this end, we propose a masked filter modeling to construct the PCA-like knowledge for local knowledge transfer, which introduces a decoder $D_{\theta_l}$ to upsample the student.

Specifically, the reconstruction loss of MFM (\emph{i.e.}, the third term in Eq.~\ref{eq_all}) is formulated as:
\begin{equation}
    \mathcal{L}_{RL}\big(\mathcal{Z}_l^{(t)}, D_{\theta_l}(\mathcal{Z}_l^{(s)}; \theta_l)\big) = \|\mathcal{Z}_l^{(t)}-D_{\theta_l}(\mathcal{Z}_l^{(s)}; \theta_l)\|_2,
\end{equation}
where $\|\cdot\|_F$ denotes the Frobenius norm.
In this paper, $D_{\theta_l}$ are built according to different CNN depths with various numbers of hidden layers, which is further validated in experiments.
In addition, instead of applying reconstruction at every sampler, the MFM is added at the end of each stage in CNNs for optimization.

\textbf{FLOPs regularization term.} Instead of L0-norm on the binary mask in Eq.~\ref{eq_gen}, we construct a FLOPs regularization term to sparsify the student network according to a pre-defined global compression rate $r$. Specifically, the FLOPs distance between the student and the teacher is minimized during training by:
\begin{equation}
    \footnotesize
    \label{frt}
    {\mathcal{R}} = \| \text{FLOPs}(\phi(\mathbf{x}_i; \mathcal{W}, \mathbf{m}(\mathbf{P}))) - (1-r) \text{FLOPs}(\phi(\mathbf{x}_i; \mathcal{W}^{(t)}))\|_F.
\end{equation}

\textbf{Knowledge distillation loss.} We also transfer the knowledge of the soft predictions from the teacher to the student (\emph{i.e.}, the second term in Eq.~\ref{eq_all}), which is formulated as:
\begin{equation}
    \label{kd}
    \mathcal{L}_{KD} = T^2 \text{KL}(\mathbf{a}_i^{(t)},\mathbf{a}_i^{(s)}),
\end{equation}
where $KL(\cdot,\cdot)$ denotes the KL divergence. $\mathbf{a}_i^{(t)}$ and $\mathbf{a}_i^{(s)}$ are the soft prediction from the teacher and the student, respectively. Specifically, they are defined as:
\begin{equation}
    \label{soft}
    \begin{split}
        & \mathbf{a}_i^{(t)} = \text{softmax}\big(\phi(\mathbf{x}_i;\mathcal{W}^{(t)})/T\big), \\
        & \mathbf{a}_i^{(s)} = \text{softmax}\big(\phi(\mathbf{x}_i;\mathcal{W},\mathbf{m}(\mathbf{P}))/T\big), \\
    \end{split}
\end{equation}
where $T$ is a softened factor, which is set to $3$ in our experiments by following \cite{yim2017gift,romero2014fitnets}.


\begin{algorithm}[t]
    \small
    \caption{The optimization of Eq.~\ref{eq_all}}
    \renewcommand{\algorithmicrequire}{\textbf{Input:}}
    \renewcommand{\algorithmicensure}{\textbf{Output:}}
    \begin{algorithmic}[1]
        \REQUIRE
        Training data $\mathcal{D} = \{\mathbf{x}_i, y_i\}_{i=1}^{N}$,
        weight set $\mathcal{W}=\{\mathcal{W}_1^1,\cdots,\mathcal{W}_L^{C_L}\}$,
        mask set $\mathcal{M}=\{\mathbf{m}_1, \cdots \mathbf{m}_L\}$ with the corresponding sampler parameter set $\mathcal{P}=\{\mathbf{P}_1, \cdots \mathbf{P}_L\}$,
        global compression rate $r$, decoder $D_\theta(\cdot)$,
        teacher with pre-trained weight $\phi(\mathcal{W}^{(t)})$, teacher FLOPs $\psi(\mathcal{W}^{(t)})$, total epoch number $E$, Fine-tuning epoch number $E_{ft}$, learning rate $\eta$,
        initial temperature $\tau_0$, final temperature $\tau_E$, soften factor $T$. \\
        \ENSURE
        Updated weights $\mathcal{W}$, decoder parameter $\theta$ and sampler parameter $\mathcal{P}$. \\
        \STATE
        Initialize $\mathcal{W}$, $\theta, \mathcal{P}$, and $e=0$. \\
        \REPEAT
        \STATE
        \textbf{Forward Pass:} \\
        \begin{itemize}
            \item update $e:=e+1$ and adjust temperature $\tau(e)$ via annealing schedule.
            \item Compute the binary mask $\mathbf{m}_l$ via Eqs.~\ref{eq7} and~\ref{eq8} and the sampler outputs $\mathcal{Z}_l^{(s)}$ via Eq.~\ref{eq5} without L0-norm restriction.
            \item Compute the decoder outputs based on $\mathcal{Z}_l^{(s)}$ via $D_{\theta_l}$, and extract pre-trained feature maps $\mathcal{Z}_l^{(t)}$.
            \item Compute the soft predictions $\mathbf{a}_{(t)}$ and $\mathbf{a}_{(s)}$ via Eq.~\ref{soft}.
            \item      Compute the total loss via Eq.~\ref{eq_all}.\\
        \end{itemize}
        \STATE
        \textbf{Backward Pass:} \\
        Compute the gradient of the loss \emph{w.r.t.} the student network's filters as $\nabla\mathcal{W}$ and the decoder parameter $\nabla\theta$ by back-propagation, as well as the sampler parameters $\nabla\mathcal{P}$.\\
        \STATE
        \textbf{Update:} \\
        Use $\nabla\mathcal{W}$, $\nabla\theta$ and $\nabla\mathcal{P}$ to update $\mathcal{W}$, $\theta$ and $\mathcal{P}$ via the AdaMax updating strategy, respectively. \\
        \UNTIL{$e$ reaches the maximum epoch number $E-E_{ft}$.} \\
        \STATE
        \textbf{Fine-tune} the final compact model with the remaining $E_{ft}$ epochs by fixing the binary mask, and removing the redundant structures, whose mask values are $0$.
    \end{algorithmic}
    \label{algorithm1}
\end{algorithm}

\subsection{Optimization}
\label{opt}
To solve the optimization problem of the proposed KDFS, we introduce the AdaMax optimizer~\cite{kingma2014adam} to minimize Eq.~\ref{eq_all}, which jointly learns the network parameter $\mathcal{W}$, binary mask $\mathbf{m}$ and sampler parameter $\mathbf{P}$. Algorithm~\ref{algorithm1} presents the optimization process.
In Algorithm~\ref{algorithm1}, we finally fine-tune the pruned network by fixing the final binary mask to reduce the marginal performance degradation.

\textbf{Pruning strategy.} To keep the matched dimension of input/output channels in a residual block, the previous works~\cite{lin2019toward,lin2019towards} added the binary mask after the first two layers in each residual block while keeping the last convolutional layer of each residual block unchanged. To achieve more flexible pruning, we prune all convolutional layers in a residual block except for the convolutional layer in the shortcut connection (an extremely small number of parameters). Thus, we employ \emph{zero-padding} on the pruned feature maps after the last convolutional layer of each residual block according to the indices of the binary mask to implement deep pruning. This simple strategy allows more flexible pruning and achieves a higher compression rate.

\section{Experiments}
\label{exp}

\begin{table*}[t]
    \caption{Results for pruning ResNet-56 on CIFAR-10. The base ResNet-56 has $0.12$G FLOPs and $0.8$M parameters. $\bold{value}$ with bold font means the best performance in all papers.}
    \small
    \centering
    \resizebox{0.9\linewidth}{!}{
        \begin{tabular}{c|cccccc}
            \toprule
            Method                                                    & \makecell{Top-1 Acc.                                                                                                              \\ {base $\rightarrow$ pruned} (\%)}                & \makecell{Top-1 Acc.                                                                    \\ {$\downarrow$} (\%)} & FLOPs & \makecell{FLOPs\\{$\downarrow$} (\%)} & Params & \makecell{Params\\{$\downarrow$} (\%)} \\
            \midrule
            (ICLR'17) L1~\cite{li2017pruning}                         & $93.04 \rightarrow 93.06$                   & -$0.02$        & $90.90$M        & $27.60$        & $0.83$M        & $13.70$        \\
            (CVPR'18) NISP~\cite{yu2018nisp}                          & $93.26 \rightarrow 93.01$                   & $0.25$         & $81.00$M        & $35.50$        & $0.49$M        & $42.40$        \\
            (CVPR'19) GAL~\cite{lin2019towards}                       & $93.26 \rightarrow 92.98$                   & $0.28$         & $78.30$M        & $37.60$        & $0.75$M        & $11.80$        \\
            (CVPR'20) HRank~\cite{lin2020hrank}                       & $93.26 \rightarrow 93.52$                   & -$0.26$        & $88.72$M        & $29.30$        & $0.71$M        & $16.80$        \\
            (CVPR'20) LeGR~\cite{chin2020towards}                     & $93.90 \rightarrow 94.10$                   & -$0.20$        & $87.80$M        & $30.00$        & $-$            & $-$            \\
            (TNNLS'21) FilterSketch~\cite{lin2021filter}              & $93.26 \rightarrow 93.65$                   & -$0.39$        & $88.05$M        & $30.43$        & $0.68$M        & $20.60$        \\
            KDFS-0.4 (ours)                                           & \textbf{93.26} $\rightarrow$ \textbf{93.78} & \textbf{-0.52} & \textbf{74.22M} & \textbf{40.85} & \textbf{0.50M} & \textbf{41.70} \\
            \midrule
            (ICCV'17) Lasso~\cite{he2017channel}                      & $93.26 \rightarrow 90.80$                   & $2.46$         & $62.00$M        & $50.06$        & $-$            & $-$            \\
            (ECCV'18) AMC~\cite{he2018amc}                            & $92.80 \rightarrow 91.90$                   & $0.90$         & $63.01$M        & $50.00$        & $-$            & $-$            \\
            (CVPR'19) ENC-Inf~\cite{kim2019efficient}                 & $93.10 \rightarrow 93.00$                   & $0.10$         & $63.01$M        & $50.00$        & $-$            & $-$            \\
            (NeurIPS'20) Zhuang \emph{et al.}~\cite{zhuang2020neuron} & $93.80 \rightarrow 93.83$                   & -$0.03$        & $-$             & $47.00$        & $-$            & $-$            \\
            (ECCV'20) DSA~\cite{ning2020dsa}                          & $93.12 \rightarrow 92.91$                   & $0.22$         & $-$             & $49.70$        & $-$            & $-$            \\
            (CVPR'20) HRank~\cite{lin2020hrank}                       & $93.26 \rightarrow 93.17$                   & $0.09$         & $63.01$M        & $50.00$        & $0.49$M        & $42.40$        \\
            (CVPR'20) Hinge~\cite{li2020group}                        & $93.69 \rightarrow 92.95$                   & $0.74$         & $63.01$M        & $50.00$        & $-$            & $-$            \\
            (CVPR'20) DMC~\cite{gao2020discrete}                      & $93.62 \rightarrow 93.69$                   & -$0.07$        & $-$             & $50.00$        & $-$            & $-$            \\
            (TNNLS'21) FilterSketch~\cite{lin2021filter}              & $93.26 \rightarrow 93.19$                   & $0.07$         & $73.36$M        & $41.50$        & $0.50$M        & $41.20$        \\
            (CVPR'21) NPPM~\cite{gao2021network}                      & $93.59 \rightarrow 93.40$                   & $0.19$         & $-$             & $50.00$        & $-$            & $-$            \\
            (ECCV'22) DDNP~\cite{gao2022disentangled}                 & $93.62 \rightarrow 93.83$                   & -$0.21$        & $-$             & $51.00$        & $-$            & $-$            \\
            KDFS-0.5 (ours)                                           & \textbf{93.26} $\rightarrow$ \textbf{93.58} & \textbf{-0.32} & \textbf{61.25M} & \textbf{51.19} & \textbf{0.41M} & \textbf{51.33} \\
            \midrule
            (IJCAI'18) SFP~\cite{he2018soft}                          & $93.59 \rightarrow 93.26$                   & $1.33$         & $-$             & $52.60$        & $-$            & $-$            \\
            (NeurIPS'19) TAS~\cite{dong2019network}                   & $-$                                         & $0.77$         & $59.40$M        & $52.60$        & $-$            & $-$            \\
            (CVPR'19) FPGM~\cite{he2019filter}                        & $93.59 \rightarrow 92.93$                   & $0.66$         & $-$             & $53.60$        & $-$            & $-$            \\
            (CVPR'19) GAL~\cite{lin2019towards}                       & $93.26 \rightarrow 90.36$                   & $2.90$         & $-$             & \textbf{60.20} & $-$            & $-$            \\
            (ICML'20) SCP~\cite{kang2020operation}                    & $93.59 \rightarrow 93.23$                   & $0.36$         & $-$             & $51.50$        & $-$            & $-$            \\
            (CVPR'20) LeGR~\cite{chin2020towards}                     & $93.90 \rightarrow 93.70$                   & $0.20$         & $58.90$M        & $53.00$        & $-$            & $-$            \\
            (NeurIPS'20) SCOP~\cite{tang2020scop}                     & $93.70 \rightarrow 93.64$                   & $0.06$         & $-$             & $56.00$        & $-$            & $56.3$         \\
            (ICLR'21) CLR~\cite{le2021network}                        & $93.59 \rightarrow 92.78$                   & $0.81$         & $59.40$M        & $52.60$        & $-$            & $-$            \\
            (TCYB'21) LRMF~\cite{zhang2021filter}                     & $93.59 \rightarrow 93.25$                   & $0.34$         & $59.40$M        & $52.60$        & $-$            & $-$            \\
            (AAAI'21) DPFPS~\cite{ruan2021dpfps}                      & $93.81 \rightarrow 93.20$                   & $0.61$         & $-$             & $52.86$        & $-$            & $46.84$        \\
            (MIR'23) MaskSparsity~\cite{jiang2023pruning}             & $94.50 \rightarrow 94.19$                   & $0.31$         & $-$             & $54.88$        & $-$            & $-$            \\
            KDFS-0.6 (ours)                                           & \textbf{93.26} $\rightarrow$ \textbf{93.19} & \textbf{0.07}  & $51.24$M        & $59.17$        & \textbf{0.33M} & \textbf{60.90} \\
            \bottomrule
        \end{tabular}
    }
    \label{table1}
\end{table*}

\begin{table}[htbp]
    \caption{Results for pruning ResNet-56 on CIFAR-100.}
    \centering
    \resizebox{\linewidth}{!}{
        \begin{tabular}{c|ccc}
            \toprule
            Method                                                    & \makecell{Top-1 Acc.                                                          \\ {base $\rightarrow$ pruned} (\%)}                & \makecell{Top-1 Acc.                 \\ {$\downarrow$} (\%)}  & \makecell{FLOPs\\{$\downarrow$} (\%)}  \\
            \midrule
            (IJCAI'18) SFP~\cite{he2018soft}                          & $-$                                         & $2.61$         & $51.60$        \\
            (CVPR'19) FPGM~\cite{he2019filter}                        & $-$                                         & $1.75$         & \textbf{52.60} \\
            (NeurIPS'20) Zhuang \emph{et al.}~\cite{zhuang2020neuron} & $72.49 \rightarrow 72.46$                   & $0.06$         & $25$           \\
            (CVPR'20) LFPC~\cite{he2020learning}                      & $-$                                         & $0.58$         & $38.01$        \\
            (TCYB'21) LRMF~\cite{zhang2021filter}                     & $69.67 \rightarrow 69.23$                   & $0.44$         & \textbf{52.60} \\
            KDFS-0.5 (ours)                                           & \textbf{71.33} $\rightarrow$ \textbf{71.65} & \textbf{-0.32} & $51.98$        \\
            \bottomrule
        \end{tabular}
    }
    \label{table2}
\end{table}

\begin{table*}[t]
    \caption{Results for pruning ResNet-110 on CIFAR-10. The base ResNet-110 has $0.25$G FLOPs and $1.7$M parameters.}
    \centering
    \resizebox{0.9\linewidth}{!}{
        \begin{tabular}{c|cccccc}
            \toprule
            Method                                    & \makecell{Top-1 Acc.                                                                                                               \\ {base $\rightarrow$ pruned} (\%)}                & \makecell{Top-1 Acc.                                                                    \\ {$\downarrow$} (\%)} & FLOPs & \makecell{FLOPs\\{$\downarrow$} (\%)} & Params & \makecell{Params\\{$\downarrow$} (\%)} \\
            \midrule
            (ICLR'17) L1~\cite{li2017pruning}         & $93.55 \rightarrow 93.30$                   & $0.25$         & $155.00$M        & $38.70$        & $1.16$M        & $32.60$        \\
            (ICLR'18) Rethink~\cite{ye2018rethinking} & $93.55 \rightarrow 93.30$                   & $0.25$         & $-$              & $40.80$        & $-$            & $-$            \\
            (IJCAI'18) SFP~\cite{he2018soft}          & $93.68 \rightarrow 93.38$                   & $0.30$         & $-$              & $40.80$        & $-$            & $-$            \\
            (CVPR'18) NISP~\cite{yu2018nisp}          & $-$                                         & $0.18$         & $-$              & $43.80$        & $-$            & $-$            \\
            (CVPR'19) GAL~\cite{lin2019towards}       & $93.39 \rightarrow 92.74$                   & $0.65$         & $-$              & $48.50$        & $0.95$M        & $44.80$        \\
            KDFS-0.5 (ours)                           & \textbf{93.50} $\rightarrow$ \textbf{94.23} & \textbf{-0.73} & \textbf{122.61M} & \textbf{51.52} & \textbf{0.81M} & \textbf{52.94} \\
            \midrule
            (IJCAI'18) SFP~\cite{he2018soft}          & $93.68 \rightarrow 92.90$                   & $0.78$         & $121$M           & $52.30$        & $-$            & $-$            \\
            (CVPR'19) FPGM~\cite{he2019filter}        & $93.68 \rightarrow 93.74$                   & -$0.06$        & $121$M           & $52.30$        & $-$            & $-$            \\
            (NeurIPS'19) TAS~\cite{dong2019network}   & $-$                                         & $0.64$         & $119$M           & $53.00$        & $-$            & $-$            \\
            (CVPR'20) HRank~\cite{lin2020hrank}       & $93.50 \rightarrow 93.36$                   & $0.14$         & $105.7$M         & $58.20$        & $0.70$M        & $59.20$        \\
            (CVPR'20) LFPC~\cite{he2020learning}      & $93.50 \rightarrow 93.07$                   & $0.43$         & $101$M           & $60.30$        & $-$            & $-$            \\
            (ICLR'21) CLR~\cite{le2021network}        & $93.68 \rightarrow 92.91$                   & $0.77$         & $121$M           & $52.30$        & $-$            & $-$            \\
            KDFS-0.6 (ours)                           & \textbf{93.50} $\rightarrow$ \textbf{93.65} & \textbf{-0.15} & \textbf{98.80M}  & \textbf{60.93} & \textbf{0.64M} & \textbf{62.87} \\
            \bottomrule
        \end{tabular}
        \label{table3}
    }
\end{table*}

\subsection{Experimental Settings}

\textbf{Datasets and base models.}
We evaluate the proposed approach on widely-used datasets and models for classification on CIFAR-10/100~\cite{krizhevsky2009learning} and ImageNet ILSVRC 2012~\cite{russakovsky2015imagenet}. Base models contain ResNets~\cite{he2016deep}, MobileNetV2~\cite{sandler2018mobilenetv2}, GoogLeNet~\cite{szegedy2015going} and DenseNet~\cite{huang2017densely}. We also evaluate the transfer ability of the pruned models using Faster-RCNN~\cite{ren2015faster} for object detection on PASCAL VOC~\cite{pascal-voc-2012} and Mask-RCNN~\cite{he2017mask} for instance segmentation on MS COCO~\cite{lin2014microsoft}.

\textbf{Implementations.}
We implement the proposed KDFS approach using PyTorch~\cite{paszke2019pytorch}.
The optimization problem defined in Eq.~\ref{eq_all} is solved using the AdaMax optimizer~\cite{kingma2014adam}. On CIFAR-10/100, we train models for $350$ epochs with an initial learning rate of $1e-2$ and employ a cosine annealing learning rate decay strategy to decay the learning rate to $1e-4$. ResNet and DenseNet are trained on a single NVIDIA RTX 3080ti GPU with a batch size of $256$, using weight decay values of $1e-4$ and $1e-5$, respectively. For GoogLeNet, we train the models on two NVIDIA RTX 3080ti GPUs with a batch size of $128$ and a weight decay of $1e-4$. On ImageNet ILSVRC 2012, we train models for $250$ epochs on four NVIDIA RTX 3080ti GPUs. The training is performed with a batch size of $256$, an initial learning rate of $4e-3$, a weight decay $2e-5$, and a cosine annealing learning rate decay strategy that decayed the learning rate to $4e-5$. During the fine-tuning, the learning rate for training all models is set to a $100$x reduction. and The pruned models are fine-tuned for $50$ epochs ($E_{ft} = 50$) on CIFAR-10/100 and $20$ epochs ($E_{ft} = 20$) on ImageNet ILSVRC 2012.
Other hyper-parameters remain consistent with the settings mentioned above.

For the balance hyper-parameters of the total objective function (Eq.~\ref{eq_all}), we set $\lambda_1$, $\lambda_2$, and $\lambda_3$ to $0.05$, $1,000$, and $10,000$, respectively. The softened factor $T$ in the knowledge distillation loss (Eq.~\ref{kd}) is set to $3$. The temperature $\tau$ in Eq.~\ref{eq7} is initialized as $\tau_0 = 1.0$ and gradually annealed to $\tau_E = 0.1$.
The global compression rate $r$ is chosen mutually such that the pruned model achieves the desired FLOPs reduction. For example, as shown in Table~\ref{table1}, we set $r$ to $0.4$, $0.5$, and $0.6$ for experiments with ResNet-56 on CIFAR-10, resulting in FLOPs reductions of $40.85\%$, $51.19\%$, and $59.17\%$, respectively.
The specific implementations for object detection and instance segmentation are discussed in Subsection~\ref{other tasks}.

\textbf{Evaluation metrics.} We use the thop package~\footnote{https://github.com/Lyken17/pytorch-OpCounter/blob/master/thop} for FLoating Point operations~(FLOPs) and Parameters~(Params) calculation.
FLOPs and Params are used as the evaluation metrics for the storage and computation cost, respectively.
We report Top-1 accuracy on CIFAR-10/100, and Top-1/5 accuracy on ImageNet.
For object detection on PASCAL VOC and instance segmentation on MS COCO, we report Average Precision~(AP).

\textbf{State-of-the-art methods.}
The base models are accessed by Pytorch model zoo~\footnote{https://pytorch.org/vision/stable/models.html} or trained from scratch using the same learning strategy and hyper-parameters mentioned above. We also make our KDFS with several state-of-the-art methods, including L1~\cite{li2017pruning}, HRank~\cite{lin2020hrank},
LeGR~\cite{chin2020towards}, FilterSketch~\cite{lin2021filter}, NISP~\cite{yu2018nisp}, GAL~\cite{lin2019towards}, Zhuang~\emph{et al.}~\cite{zhuang2020neuron},
DSA~\cite{ning2020dsa}, NPPM~\cite{gao2021network}, ENC-Inf~\cite{kim2019efficient}, AMC~\cite{he2018amc}, Hinge~\cite{li2020group}, DMC~\cite{gao2020discrete},
Lasso~\cite{he2017channel}, DDNP~\cite{gao2022disentangled}, SCP~\cite{kang2020operation}, SFP~\cite{he2018soft}, CLR~\cite{le2021network}, LRMF~\cite{zhang2021filter},
TAS~\cite{dong2019network}, DPFPS~\cite{ruan2021dpfps}, FPGM~\cite{he2019filter}, MaskSparsity~\cite{jiang2023pruning}, SCOP~\cite{tang2020scop},
LFPC~\cite{he2020learning}, Rethink~\cite{ye2018rethinking}, Zhao~\emph{et al.}~\cite{zhao2019variational}, EZCrop~\cite{lin2022ezcrop}, LSTM~\cite{ding2020prune},
SRR-GR~\cite{wang2021convolutional}, DMCP~\cite{guo2020dmcp}, LRF~\cite{joo2021linearly}, HTP-URC~\cite{qian2023hierarchical}, GBN~\cite{you2019gate},
Uniform~\cite{sandler2018mobilenetv2}, MetaPruning~\cite{liu2019metapruning}, GFS~\cite{ye2020good}, Liu~\emph{et al.}~\cite{liu2021joint}, PCP~\cite{guo2020model} and DNCP~\cite{zheng2022model}.

\textbf{Feature reconstruction networks.} 
We select two kinds of CNNs with different depths, \emph{i.e.}, a one-hidden layer or a two-hidden layer to explore their effect. For the neuron number of the $l$-th layer in the decoder, we simply set the same filter number to the row dimension (\emph{i.e.}, $C_l$) of the sampler. Additionally, we only use the reconstruction loss after the last layer of each stage in CNNs.

\subsection{Comparison with the State-of-the-art Methods.}

\subsubsection{CIFAR-10 and CIFAR-100}
We evaluate the performance on CIFAR-10/100 with several popular CNNs, including ResNet-56/110, DenseNet-40, and GoogLeNet. DenseNet-40 contains $40$ layers with a growth rate of $12$. For GoogLeNet, the dimension of the fully-connected layer in the original GoogLeNet is changed to the class number for CIFAR-10/100.

\begin{figure*}[t]
    \centering
    \includegraphics[scale = 0.2]{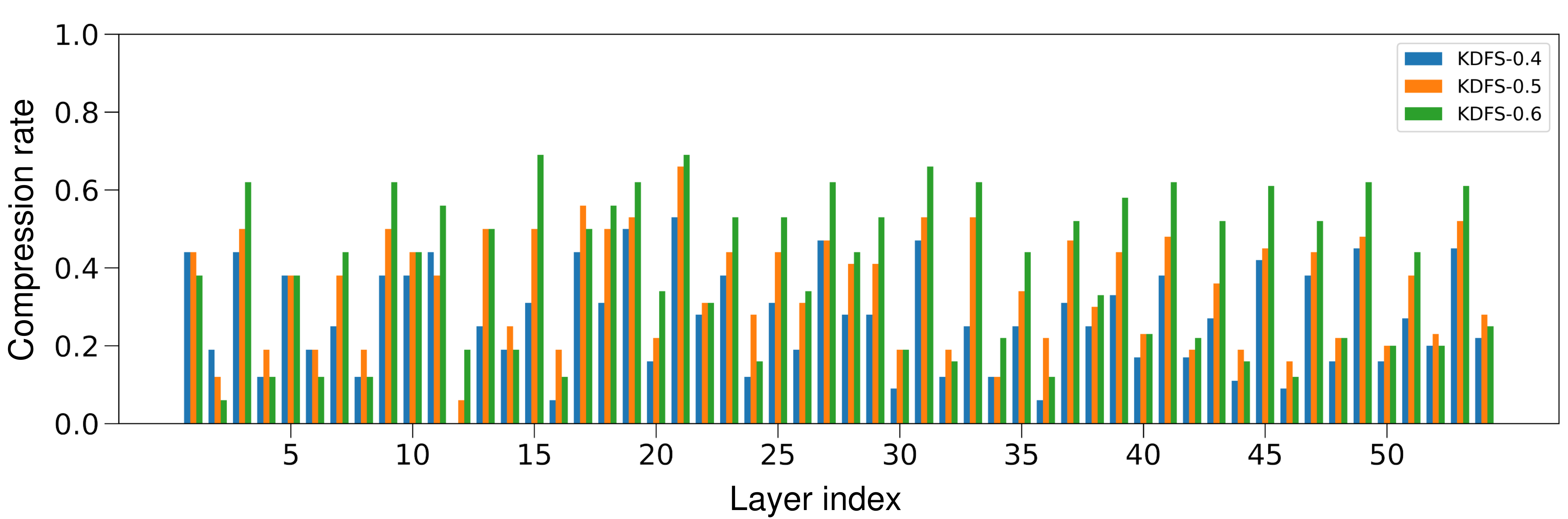}
    \caption{Layer-wise compression rates of pruned ResNet-56 with a total $27$ blocks (\emph{i.e.}, $54$ layers) on CIFAR-10. We do not prune the first convolution and the last fully-connected layer for dimension matching on ResNet-56. The layer indices $[1,3,5,\cdots,53]$ refer to the first layer in each block.}
    \label{fig6}
\end{figure*}

\textbf{ResNets compression.}
Table~\ref{table1} summarizes the experimental results for pruning ResNet-56 on CIFAR-10. We apply KDFS at three FLOPs reduction levels by setting the global compression rate $r$ to $0.4$, $0.5$, and $0.6$.
Our KDFS-0.4 and KDFS-0.5 reduce $40.85\%$ and $51.19\%$ FLOPs, while Top-1 accuracy is increased by $0.52$ and $0.32$, respectively. Furthermore, KDFS-0.6 achieves $59.17\%$ FLOPs reduction, while its Top-1 accuracy only drops $0.07$. At all three FLOPs reduction levels, the proposed KDFS achieves the best performance compared to previous SOTA methods. For example, compared to FilterSketch and DDNP, KDFS-0.4 and KDFS-0.5 achieves the highest accuracy improvement over the base model ($0.52$ \emph{vs.} $0.39$ in FilterSketch, and $0.32$ \emph{vs.} $0.21$ in DDNP) with the highest FLOPs reduction ($40.85\%$ \emph{vs.} $30.43\%$ in FilterSketch, and $51.19\%$ \emph{vs.} $51.0\%$ in DDNP).
When more filters are pruned, KDFS-0.6 achieves the higher pruned rate in FLOPs ($59.17\%$ \emph{vs.} $56.0\%$) and Params ($60.90\%$ \emph{vs.} $56.3\%$) with almost the same accuracy drop, compared to SCOP. We also visualize the pruning rate of different layers by setting $r$ to $0.4$, $0.5$, and $0.6$, as shown in Fig.~\ref{fig6}.
Our method tends to prune the first layer of each block, which is more flexible than the second one. This is because, under the constraint of FLOPs, KDFS tends to prune structures that contribute a larger portion of the FLOPs. When the first layer of each block is pruned, it reduces the FLOPs of the first layer and the FLOPs of the second layer.


\begin{table}[t]
    \caption{Results for pruning DenseNet-40 on CIFAR-10. The base DenseNet-40 has $0.28$G FLOPs and $1.0$M parameters.}
    \centering
    \resizebox{\linewidth}{!}{
        \begin{tabular}{c|ccc}
            \toprule
            Method                                                  & \makecell{Top-1 Acc.                                                         \\ {base $\rightarrow$ pruned} (\%)}                & \makecell{Top-1 Acc.                                                                    \\ {$\downarrow$} (\%)}  & \makecell{FLOPs\\{$\downarrow$} (\%)}  \\
            \midrule
            (CVPR'19) Zhao \emph{et al.}~\cite{zhao2019variational} & $94.81 \rightarrow 93.16$                   & $1.65$        & $44.8$         \\
            (CVPR'19) GAL~\cite{lin2019towards}                     & $94.81 \rightarrow 93.53$                   & $1.28$        & $54.7$         \\
            (CVPR'20) HRank~\cite{lin2020hrank}                     & $94.81 \rightarrow 93.68$                   & $1.13$        & $61.0$         \\
            (WACV'22) EZCrop~\cite{lin2022ezcrop}                   & $94.81 \rightarrow 93.76$                   & $1.05$        & $59.9$         \\
            KDFS-0.6 (ours)                                         & \textbf{94.81} $\rightarrow$ \textbf{94.44} & \textbf{0.37} & \textbf{61.78} \\
            \bottomrule
        \end{tabular}
    }
    \label{table4}
\end{table}

\begin{figure}[t]
    \centering
    \includegraphics[scale = 0.18]{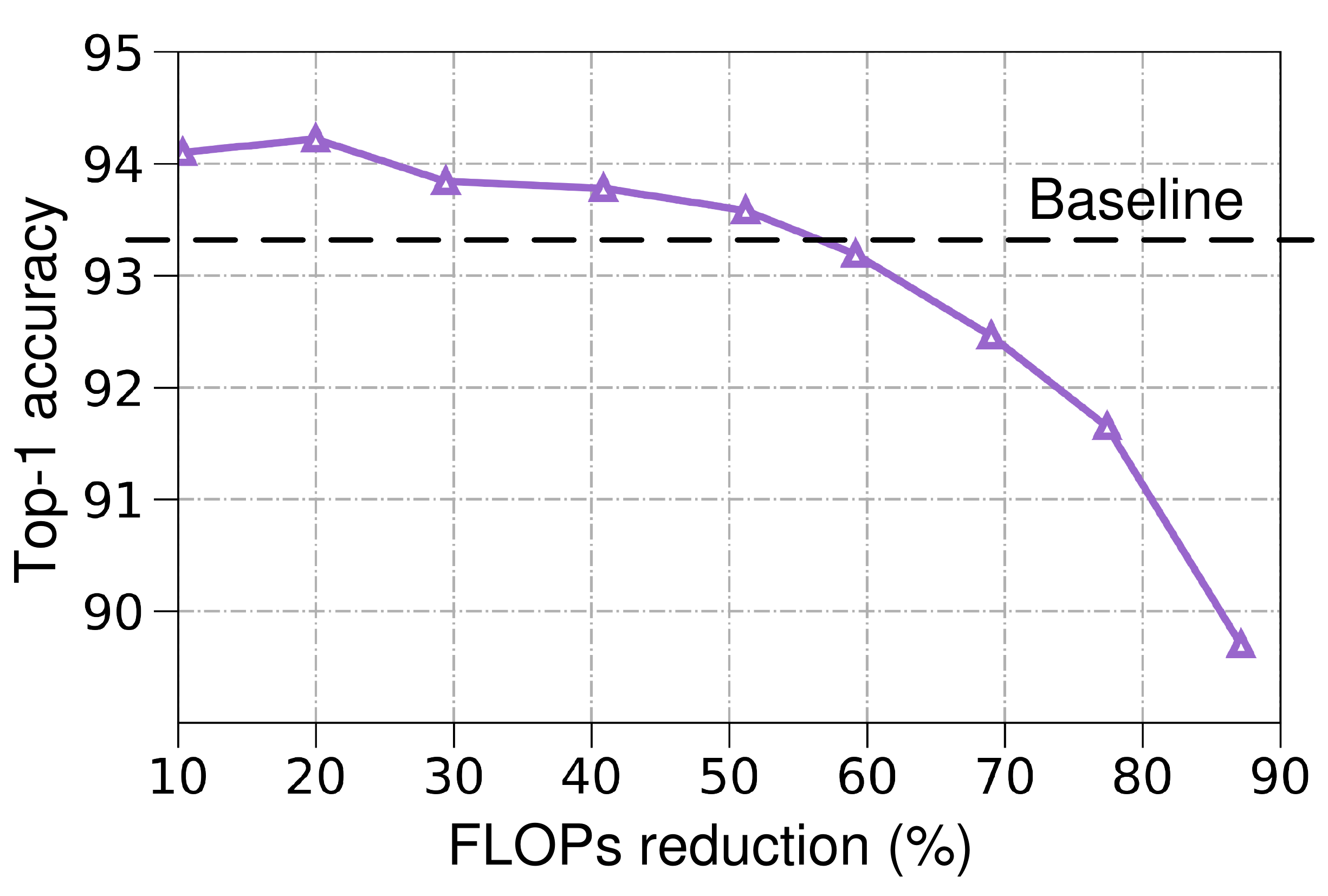}
    \caption{Analysis of the varied FLOPs reductions for pruning ResNet-56 on CIFAR-10.}
    \label{fig4}
\end{figure}

In addition, we vary $r$ from $0.1$ to $0.9$ to obtain the pruning trend, as shown in Fig.~\ref{fig4}. When $r$ is smaller than $0.6$, our method achieves higher accuracies than base ResNet-56. When the value of $r$ surpasses $0.6$, the accuracy of our method drops, especially the large gap between the results of $0.8$ and $0.9$.

For CIFAR-100, the experimental results for compressing ResNet-56 are shown in Table~\ref{table2}.
Compared to Zhuang~\emph{et al.}, LFPC, SFP, FPGM, and LRMF, our KDFS-0.5 achieves the best trade-off between accuracy improvement and FLOPs reduction. Compared to LRMF, with a slightly lower FLOPs reduction ($51.98\%$ \emph{vs.} $52.60\%$), our KDFS achieves significant accuracy improvement over LRMF by $0.76\%$ (\emph{i.e.}, $0.34\%$ \emph{vs.} $-0.44\%$). Our method also outperforms Zhuang~\emph{et al.} by $0.48\%$ and $26.98\%$ regarding accuracy improvement and FLOPs reduction.

\begin{table}[t]
    \caption{Results for pruning GoogLeNet on CIFAR-10. The base GoogLeNet has $1.5$G FLOPs and $6.1$M parameters.}
    \centering
    \resizebox{\linewidth}{!}{
        \begin{tabular}{c|ccc}
            \toprule
            Method                                       & \makecell{Top-1 Acc.                                                          \\ {base $\rightarrow$ pruned} (\%)}                & \makecell{Top-1 Acc.                                                                    \\ {$\downarrow$} (\%)}  & \makecell{FLOPs\\{$\downarrow$} (\%)}  \\
            \midrule
            (ICLR'17) L1~\cite{li2017pruning}            & $95.05 \rightarrow 94.54$                   & $0.51$         & $32.9$         \\
            (CVPR'19) GAL~\cite{lin2019towards}          & $95.05 \rightarrow 93.93$                   & $1.12$         & $38.2$         \\
            (CVPR'20) HRank~\cite{lin2020hrank}          & $95.05 \rightarrow 94.53$                   & $0.52$         & $54.9$         \\
            (TNNLS'21) FilterSketch~\cite{lin2021filter} & $95.05 \rightarrow 94.88$                   & $0.17$         & $61.1$         \\
            KDFS-0.6 (ours)                              & \textbf{95.05} $\rightarrow$ \textbf{95.24} & \textbf{-0.19} & \textbf{62.12} \\
            \bottomrule
        \end{tabular}
    }
    \label{table5}
\end{table}

\begin{table*}[t]
    \caption{Results for pruning ResNet-50 on ImageNet. The base ResNet-50 has $4.1$G FLOPs and $25.6$M parameters.}
    \centering
    \resizebox{\linewidth}{!}{
        \begin{tabular}{c|cccccc}
            \toprule
            Method                                                    & \makecell{Top-1 Acc.                                                                                                                                          \\ {base $\rightarrow$ pruned} (\%)}                & \makecell{Top-1 Acc.                                                                                   \\ {$\downarrow$} (\%)} & \makecell{Top-5 Acc.                                                          \\ {base $\rightarrow$ pruned} (\%)} & \makecell{Top-5 Acc.\\ {$\downarrow$} (\%)} & \makecell{FLOPs\\{$\downarrow$} (\%)} & \makecell{Params\\{$\downarrow$} (\%)} \\
            \midrule
            (ICCV'17) Lasso~\cite{he2017channel}                      & $76.15 \rightarrow 72.30$                   & $3.85$         & $92.96 \rightarrow 90.80$                   & $2.16$         & $34.10$        & $-$            \\
            (IJCAI'18) SFP~\cite{he2018soft}                          & $76.15 \rightarrow 74.61$                   & $1.54$         & $92.87 \rightarrow 92.06$                   & $0.81$         & $41.80$        & $-$            \\
            (TNNLS'21) FilterSketch~\cite{lin2021filter}              & $76.13 \rightarrow 75.22$                   & $0.91$         & $92.86 \rightarrow 92.41$                   & $0.45$         & $35.50$        & $33.50$        \\
            (ICLR'21) CLR~\cite{le2021network}                        & $76.01 \rightarrow 74.85$                   & $1.16$         & $92.96 \rightarrow 92.31$                   & $0.45$         & $40.39$        & $33.80$        \\
            (TNNLS'22) DNCP~\cite{zheng2022model}                     & $76.60 \rightarrow 76.30$                   & $0.30$         & $-$                                         & $-$            & \textbf{46.34} & $-$            \\
            KDFS-0.4 (ours)                                           & \textbf{76.15} $\rightarrow$ \textbf{76.26} & \textbf{-0.11} & \textbf{92.87} $\rightarrow$ \textbf{93.07} & \textbf{-0.20} & $42.32$        & \textbf{34.11} \\
            \midrule
            (CVPR'19) GAL~\cite{lin2019towards}                       & $76.15 \rightarrow 71.95$                   & $4.20$         & $92.87 \rightarrow 90.94$                   & $1.93$         & $43.03$        & $-$            \\
            (CVPR'19) MetaPruning~\cite{liu2019metapruning}           & $76.60 \rightarrow 75.40$                   & $1.20$         & $-$                                         & $-$            & $51.2$         & $-$            \\
            (NeurIPS'19) GBN~\cite{you2019gate}                       & $75.85 \rightarrow 75.18$                   & $0.67$         & $92.67 \rightarrow 92.41$                   & $0.26$         & $55.06$        & \textbf{53.40} \\
            (TIP'20) LSTM~\cite{ding2020prune}                        & $76.12 \rightarrow 75.00$                   & $1.12$         & $93.00 \rightarrow 92.67$                   & $0.33$         & $43.00$        & $-$            \\
            (CVPR'20) HRank~\cite{lin2020hrank}                       & $76.15 \rightarrow 74.98$                   & $1.17$         & $92.87 \rightarrow 92.33$                   & $0.54$         & $43.76$        & $-$            \\
            (CVPR'20) DMCP~\cite{guo2020dmcp}                         & $76.6 \rightarrow 76.2$                     & $0.4$          & $-$                                         & $-$            & $46.21$        & $-$            \\
            (NeurIPS'20) Zhuang \emph{et al.}~\cite{zhuang2020neuron} & $76.15 \rightarrow 75.63$                   & $0.52$         & $-$                                         & $-$            & $54$           & $-$            \\
            (ICML'20) SCP~\cite{kang2020operation}                    & $75.89 \rightarrow 75.27$                   & $0.62$         & $92.86 \rightarrow 92.18$                   & $0.68$         & $54.3$         & $-$            \\
            (NeurIPS'20) SCOP~\cite{tang2020scop}                     & $76.15 \rightarrow 75.26$                   & $0.89$         & $92.87 \rightarrow 92.53$                   & $0.34$         & $54.6$         & $51.8$         \\
            (CVPR'20) DMC~\cite{gao2020discrete}                      & $76.15 \rightarrow 75.35$                   & $0.80$         & $92.87 \rightarrow 92.49$                   & $0.38$         & $55.0$         & $-$            \\
            (TCSVT'20) PCP~\cite{guo2020model}                        & $74.3 \rightarrow 73.4$                     & $0.9$          & $92.1 \rightarrow 91.5$                     & $0.6$          & \textbf{55.5}  & $40.8$         \\
            (CVPR'21) SRR-GR~\cite{wang2021convolutional}             & $76.13 \rightarrow 75.76$                   & $0.37$         & $92.86 \rightarrow 92.67$                   & $0.19$         & $44.1$         & $-$            \\
            (TNNLS'21) FilterSketch~\cite{lin2021filter}              & $76.13 \rightarrow 74.68$                   & $1.45$         & $92.86 \rightarrow 92.17$                   & $0.69$         & $45.50$        & $43.00$        \\
            (AAAI'21) DPFPS~\cite{ruan2021dpfps}                      & $76.15 \rightarrow 75.55$                   & $0.60$         & $92.87 \rightarrow 92.54$                   & $0.33$         & $46.20$        & $-$            \\
            (TIP'21) Liu \emph{et al.}~\cite{liu2021joint}            & $76.60 \rightarrow 76.00$                     & $0.6$          & $-$                                         & $-$            & $53.7$         & $-$            \\
            (AAAI'21) LRF~\cite{joo2021linearly}                      & $76.21 \rightarrow 75.78$                   & $0.43$         & \textbf{92.88} $\rightarrow$ \textbf{92.85} & \textbf{0.03}  & $51.8$         & $49.1$         \\
            (CVPR'21) SRR-GR~\cite{wang2021convolutional}             & $76.13 \rightarrow 75.11$                   & $1.02$         & $92.86 \rightarrow 92.35$                   & $0.51$         & $55.1$         & $-$            \\
            (WACV'22) EZCrop~\cite{lin2022ezcrop}                     & $76.15 \rightarrow 75.68$                   & $0.47$         & $92.87 \rightarrow 92.70$                   & $0.17$         & $44.20$        & $40.82$        \\
            (MIR'23) MaskSparsity~\cite{jiang2023pruning}             & $76.44 \rightarrow 75.68$                   & $0.76$         & $93.22 \rightarrow 92.78$                   & $0.44$         & $51.07$        & $-$            \\
            (TNNLS'23) HTP-URC~\cite{qian2023hierarchical}            & \textbf{76.13} $\rightarrow$ \textbf{75.81} & \textbf{0.32}  & $-$                                         & $-$            & $54$           & $38$           \\
            KDFS-0.5 (ours)                                           & $76.15 \rightarrow 75.80$                   & $0.35$         & $92.87 \rightarrow 92.66$                   & $0.21$         & $55.36$        & $42.86$        \\
            \bottomrule
        \end{tabular}
    }
    \label{table6}
\end{table*}

In Table~\ref{table3}, we further compress deeper ResNet-110 on CIFAR-10. Our KDFS also achieves the best performance compared to previous SOTA methods. For example, our KDFS-0.6 is able to improve the accuracy of the base ResNet-110 by $0.15\%$ (\emph{vs.} $-0.14\%$ and $-0.43\%$) and achieves the highest $60.93\%$ (\emph{vs.} 58.20\% and 60.30\%) FLOPs reduction, compared to HRank and LFPC.

\textbf{DenseNet-40 compression.} For compressing the dense block in DenseNet-40, our method still achieves the best compression performance. As shown in Table~\ref{table4}, our KDFS-0.6 achieves the lowest accuracy drop of $0.37\%$, while reducing the highest FLOPs percentage of $61.78\%$, compared to Zhao~\emph{et al.}, GAL, EZCrop, and HRank.

\textbf{GoogLeNet compression.} Table~\ref{table5} summarizes the experimental results for compressing multi-branch GoogLeNet. We observe that our KDFS is also effective in compressing GoogLeNet, which achieves the best performance compared to other methods. Our KDFS-0.6 outperforms FilterSketch by $0.36\%$ accuracy improvement and about $1\%$ FLOPs reduction.

\subsubsection{ImageNet ILSVRC 2012}
We further evaluate the performance on ImageNet with two base models (\emph{i.e.}, ResNet-50, and MobileNetV2) to verify the effectiveness of the proposed KDFS.

\begin{table}[t]
    \caption{Results for pruning MobileNetV2 on ImageNet. The base MobileNetV2 has $0.32$G FLOPs and $3.5$M parameters.}
    \centering
    \resizebox{\linewidth}{!}{
        \begin{tabular}{c|cccc}
            \toprule
            Method                                          & \makecell{Top-1 Acc.                                                    \\ {$\downarrow$} (\%)}                & \makecell{Top-5 Acc.\\ {$\downarrow$} (\%)}  & \makecell{FLOPs\\{$\downarrow$} (\%)} &  \makecell{Params\\{$\downarrow$} (\%)}\\
            \midrule
            (CVPR'18) Uniform~\cite{sandler2018mobilenetv2} & $2.00$               & $1.40$         & $30.0$         & $-$            \\
            (ECCV'18) AMC~\cite{he2018amc}                  & $1.00$               & $-$            & $30.0$         & $-$            \\
            (CVPR'19) MetaPruning~\cite{liu2019metapruning} & $0.80$               & $-$            & $30.0$         & $-$            \\
            (ICML'20) GFS~\cite{ye2020good}                 & $0.40$               & $-$            & $30.0$         & $-$            \\
            (ECCV'22) DDNP~\cite{gao2022disentangled}       & -$0.15$              & -$0.12$        & $29.5$         & $-$            \\
            KDFS-0.3 (ours)                                 & \textbf{-0.33}       & \textbf{-0.19} & \textbf{31.95} & \textbf{27.50} \\
            \bottomrule
        \end{tabular}
    }
    \label{table7}
\end{table}

\textbf{ResNet-50 compression.}
As shown in Table~\ref{table6}, our KDFS achieves state-of-the-art performance for pruning ResNet-50 on ImageNet. By setting $r$ to $0.4$, our KDFS-0.4 achieves $0.11\%$ and $0.2\%$ Top-1 and Top-5 accuracy improvement over the base ResNet-50, while other pruning methods significantly reduce their accuracies. Furthermore, KDFS-0.4 also achieves the highest compression rate with $42.32\%$ and $34.11\%$ in terms of reducing FLOPs and parameters, respectively. When setting $r$ to $0.5$, our method achieves the best trade-off between accuracy improvement over the base model and FLOPs/parameter reduction percentage. For example, KDFS-0.5 achieves a slightly higher Top-1 accuracy drop of $0.03$ (\emph{i.e.}, $0.35$ \emph{vs.} $0.32$), while reducing significantly higher parameter percentage of $42.86\%$ (\emph{vs.} $38\%$), compared to HTP-URC. Compared to GBN, our method achieves $0.32\%$ Top-1 accuracy improvement (\emph{i.e.}, $0.35$ \emph{vs.} $0.67$) and also reduces more FLOPs reduction of $0.3\%$ (\emph{i.e.}, $55.36$ \emph{vs.} $55.06$).

\textbf{MobileNetV2 compression.} We further prune more light MobileNetV2 to verify the effectiveness of KDFS. As presented in Table~\ref{table7}, at the FLOPs reduction of about $30\%$, our KDFS-0.3 achieves the lowest accuracy drop, compared to other pruning methods. For example, our KDFS-0.3 improves the Top-1 and Top-5 accuracy of base MobileNetV2 by $0.33$ and $0.19$, which is significantly better than DDNP with $0.15\%$ and $0.12\%$ in terms of Top-1 and Top-5 accuracy, respectively.

\begin{figure}[t]
    \centering
    \includegraphics[scale = 0.25]{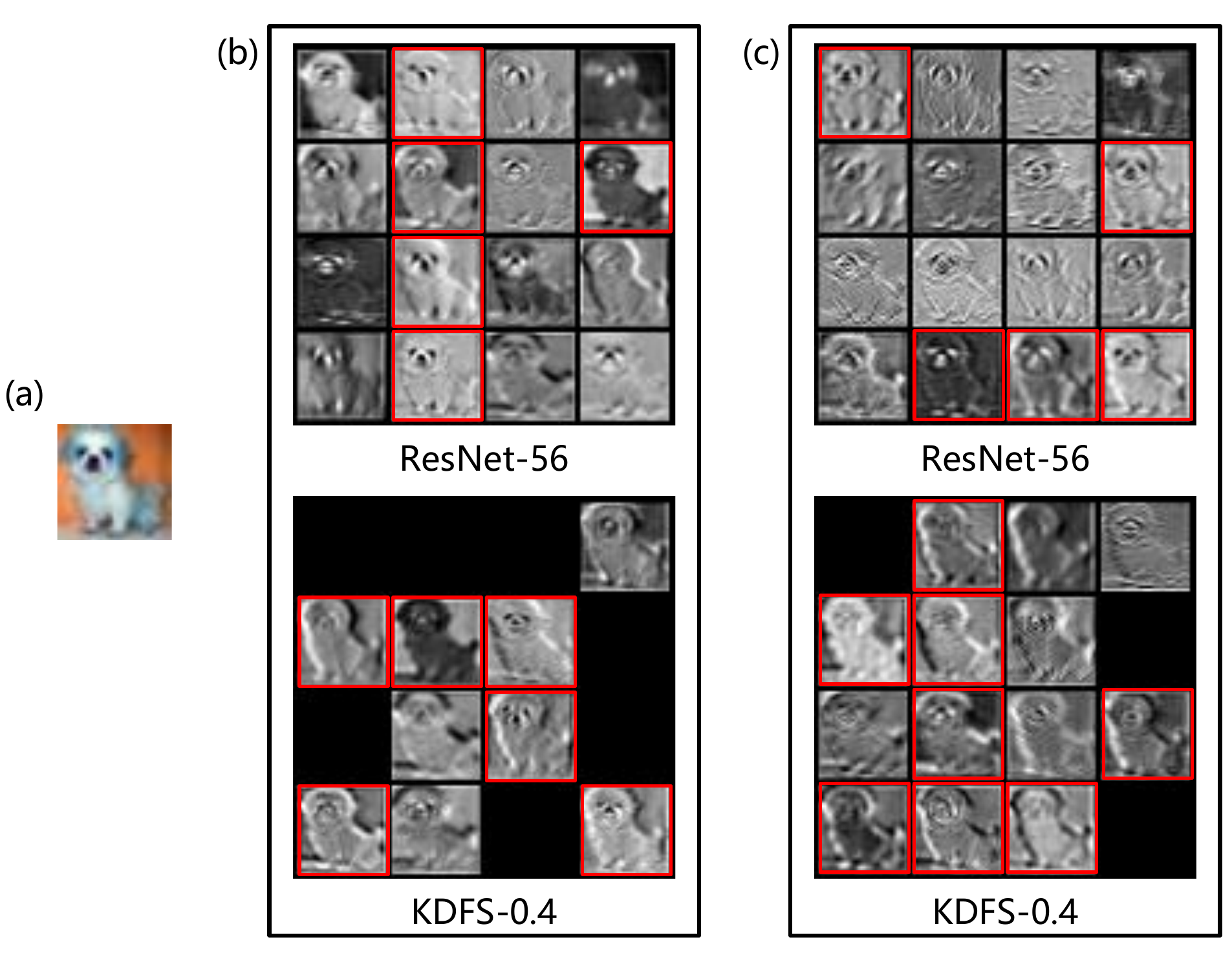}
    \caption{Feature visualization for baseline and pruned ResNet-56 on CIFAR-10. (a) Input, (b) The first layer of the first block, and (c) The second layer of the first block. The red boxes indicate that the features have a clear pattern and edge preservation.}
    \label{fig7}
\end{figure}

\subsubsection{Feature Visualization}
We visualize the feature difference between the base ResNet-56 and KDFS-0.4, which is shown in Fig.~\ref{fig7}. We visualize the features from the first and second layers at the first block, where the black boxes are masked. We observed that the pruned ResNet-56 by KDFS retains a clearer pattern and edge preservation compared to the baseline ResNet-56. It demonstrates that KDFS is able to automatically determine the important filters while enhancing the features.


\begin{table}[t]
    \caption{Effect of $\lambda$ for pruning ResNet-56 on CIFAR-10.}
    \centering
    \resizebox{1\linewidth}{!}{
        \begin{tabular}{c|c|ccc}
            \toprule
            Method   & ($\lambda_1,\lambda_2,\lambda_3$) & Top-1 Acc. (\%) & \makecell{FLOPs\\{$\downarrow$} (\%)} & \makecell{Params\\{$\downarrow$} (\%)} \\
            \midrule
                     & (0.05, 1,000, 10,000)             & $93.78$         & $40.85$                                                     & $41.70$                                                      \\
                     & (0.005, 1,000, 10,000)            & $93.76$         & $40.92$                                                     & $42.41$                                                      \\
                     & (0.5, 1000, 10,000)               & $93.48$         & $40.43$                                                     & $43.41$                                                      \\
            KDFS-0.4 & (0.05, 100, 10,000)               & $93.78$         & $40.35$                                                     & $43.66$                                                      \\
                     & (0.05, 10,000, 10,000)            & $93.12$         & $41.24$                                                     & $42.57$                                                      \\
                     & (0.05, 1,000, 1000)               & $93.60$         & $41.27$                                                     & $42.04$                                                      \\
                     & (0.05, 1,000, 100,000)            & $93.60$         & $39.29$                                                     & $38.60$                                                      \\
            \midrule
                     & (0.05, 1,000, 10,000)             & $93.58$         & $51.19$                                                     & $51.33$                                                      \\
                     & (0.005, 1,000, 10,000)            & $93.46$         & $51.57$                                                     & $51.67$                                                      \\
                     & (0.5, 1000, 10,000)               & $93.25$         & $50.79$                                                     & $53.50$                                                      \\
            KDFS-0.5 & (0.05, 100, 10,000)               & $93.38$         & $51.37$                                                     & $54.87$                                                      \\
                     & (0.05, 10,000, 10,000)            & $92.90$         & $52.03$                                                     & $53.73$                                                      \\
                     & (0.05, 1,000, 1000)               & $93.56$         & $50.84$                                                     & $52.21$                                                      \\
                     & (0.05, 1,000, 100,000)            & $93.63$         & $49.91$                                                     & $47.45$                                                      \\
            \midrule
                     & (0.05, 1,000, 10,000)             & $93.19$         & $59.17$                                                     & $60.90$                                                      \\
                     & (0.005, 1,000, 10,000)            & $93.15$         & $59.59$                                                     & $59.73$                                                      \\
                     & (0.5, 1000, 10,000)               & $92.96$         & $59.27$                                                     & $60.65$                                                      \\
            KDFS-0.6 & (0.05, 100, 10,000)               & $93.07$         & $59.07$                                                     & $61.58$                                                      \\
                     & (0.05, 10,000, 10,000)            & $92.67$         & $59.78$                                                     & $61.73$                                                      \\
                     & (0.05, 1,000, 1000)               & $93.22$         & $59.13$                                                     & $61.26$                                                      \\
                     & (0.05, 1,000, 100,000)            & $92.93$         & $58.84$                                                     & $56.57$                                                      \\
            \bottomrule
        \end{tabular}
    }
    \label{table8}
\end{table}

\subsection{Ablation Study}
In this Subsection, we evaluate the effectiveness of the differential filter sampler, and PCA-like knowledge from the decoder, which are key components of KDFS. We select ResNet-56 on CIFAR-10 for an ablation study.

\subsubsection{Hyper-parameter sensitivity of balanced weight $\lambda$ and temperature $\tau$}
First, we conduct the ablation study to verify the sensitivity of balanced weights ($\lambda_1$, $\lambda_2$, and $\lambda_3$) in Eq.~\ref{eq_all}. As presented in Table~\ref{table8}, these hyper-parameters is low sensitivity to the FLOPs and parameters, which are mainly controlled by the global compression rate $r$. Setting ($\lambda_1$, $\lambda_2$ and $\lambda_3$) to ($0.05$, $1,000$, $10,000$) achieves the best trade-off between accuracy and FLOPs/parameter reduction. Thus, we select this setting for all experiments.

We further analyze the sensitivity of temperature $\tau$ using the annealing schedule in Eq.~\ref{eq7}, shown in Table~\ref{table9}. Similar to balanced weights, the beginning and end temperatures are relatively low sensitivity to the pruning rate. However, it has a large impact on the accuracy. Setting the decay range of $\tau$ from $1$ to $0.1$ achieves the best trade-off between accuracy and FLOPs/parameter reduction, which is the default setting for all experiments.

\begin{table}[t]
    \caption{Effect of temperature $\tau$ in the annealing schedule for pruning ResNet-56 on CIFAR-10. $r$ is set to $0.6$, and $\tau_1\rightarrow\tau_2$ means the temperature varies from the beginning $\tau_1$ to the end $\tau_2$.}
    \centering
    \resizebox{0.9\linewidth}{!}{
        \begin{tabular}{c|ccc}
            \toprule
            $\tau$                 & Top-1 Acc. (\%) & \makecell{FLOPs\\{$\downarrow$} (\%)} & \makecell{Params\\{$\downarrow$} (\%)} \\
            \midrule
            $10 \rightarrow 1$     & $92.97$         & $60.68$                                                     & $60.30$                                                      \\
            $10 \rightarrow 0.1$   & $93.16$         & $59.10$                                                     & $58.88$                                                      \\
            $1 \rightarrow 1$      & $93.05$         & $59.56$                                                     & $60.28$                                                      \\
            $1 \rightarrow 0.1$    & $93.19$         & $59.17$                                                     & $60.90$                                                      \\
            $1 \rightarrow 0.01$   & $92.90$         & $59.56$                                                     & $61.49$                                                      \\
            $0.1 \rightarrow 0.1$  & $93.00$         & $60.60$                                                     & $61.18$                                                      \\
            $0.1 \rightarrow 0.01$ & $92.68$         & $59.43$                                                     & $61.60$                                                      \\
            \bottomrule
        \end{tabular}
    }
    \label{table9}
\end{table}

\begin{table}[t]
    \caption{Effect of dark knowledge and masked filter modeling for pruning ResNet-56. Zero means to use the direct alignment between the teacher and the sampler output.}
    \centering
    \resizebox{1\linewidth}{!}{
        \begin{tabular}{c|ccc|ccc}
            \toprule
            \multirow{2}{*}{DK} & \multicolumn{3}{c|}{Reconstruction loss with decoder} & \multicolumn{3}{c}{Top-1 Acc. (\%)}                                                                                                                   \\
            \cmidrule{2-7}
                                            & \multirow{1}{*}{Zero}              & \multirow{1}{*}{1-layer}         & \multirow{1}{*}{2-layer} & \multirow{1}{*}{KDFS-0.4} & \multirow{1}{*}{KDFS-0.5} & \multirow{1}{*}{KDFS-0.6}                         \\
            \midrule
            $\times$                        & $\times$                                 & $\times$                            & $\times$                    & $92.97$                   & $93.07$                   & $92.31$                   \\
            $\checkmark$                    & $\times$                                 & $\times$                            & $\times$                    & $93.55$                   & $93.17$                   & $92.66$                   \\
            $\times$                        & $\times$                                 & $\checkmark$                        & $\times$                    & $93.58$                   & $93.49$                   & $93.11$                   \\
            $\checkmark$                    & $\checkmark$                             & $\times$                            & $\times$                    & $93.75$                   & $93.38$                   & $93.10$                   \\
            $\checkmark$                    & $\times$                                 & $\checkmark$                        & $\times$                    & \textbf{93.78}            & \textbf{93.58}            & \textbf{93.19}            \\
            $\checkmark$                    & $\times$                                 & $\times$                            & $\checkmark$                & $93.76$                   & $93.57$                   & \textbf{93.19}            \\
            \bottomrule
        \end{tabular}
    }
    \label{table10}
\end{table}

\begin{table}[t]
    \caption{Effect of differential filter sampler for pruning ResNet-56 on CIFAR-10. Random or Scratch denotes the pruned models are trained by randomly selecting the important filters or the same important filter index as ours according to our pruned models, and their important filter index is fixed during training}
    \centering
    \resizebox{\linewidth}{!}{
        \begin{tabular}{c|ccc|cc}
            \toprule
            \multirow{3}{*}{Method} & \multicolumn{3}{c|}{Top-1 Acc. (\%)} & \multirow{3}{*}{\makecell{FLOPs\\{$\downarrow$} (\%)}} & \multirow{3}{*}{\makecell{Params\\{$\downarrow$} (\%)}}                     \\
            \cmidrule{2-4}
                                    & \multirow{2}{*}{Random}              & \multirow{2}{*}{Scratch}                                                     & \multirow{2}{*}{KDFS}                                                         &         &         \\
                                    &                                      &                                                                              &                                                                               &         &         \\
            \midrule
            KDFS-0.4                & $93.17\pm 0.05$                      & $93.32$                                                                      & \textbf{93.78}                                                                & $40.85$ & $41.70$ \\
            KDFS-0.5                & $93.08\pm 0.08$                      & $93.41$                                                                      & \textbf{93.58}                                                                & $51.19$ & $51.33$ \\
            KDFS-0.6                & $92.97\pm 0.01$                      & $93.04$                                                                      & \textbf{93.19}                                                                & $59.17$ & $60.90$ \\
            \bottomrule
        \end{tabular}
    }
    \label{table11}
\end{table}


\subsubsection{Effect of masked filter modeling}
Table~\ref{table10} presents the effect of dark knowledge distillation loss and masked filter modeling loss in Eq.~\ref{eq_all}. First, we analyze the effectiveness of dark knowledge. With the dark knowledge from the teacher of softened prediction, the accuracy of KDFS-0.4 is increased by $0.58\%$ (see the second row \emph{vs.} the first one) and $0.2\%$ (see the fifth row \emph{vs.} the third one). It indicates the effectiveness of dark knowledge. Second, the usage of masked filter modeling with a one-hidden layer achieves the highest accuracy of $93.78\%$, $93.58\%$, and $93.19\%$ (see last three rows) in KDFS-0.4, KDFS-0.5, and KDFS-0.6, respectively. It verifies the effectiveness of PCA-like knowledge and indicates that a deeper decoder cannot achieve higher performance due to the difficulty to transfer rich knowledge with the deeper decoder.
%


\begin{table*}[t]
    \caption{The transferring results of the pruned ResNet-50 backbone for Object detection on Faster-RCNN and instance segmentation on Mask-RCNN. FPS is evaluated by one 3080ti GPU.}
    \centering
    \resizebox{\linewidth}{!}{
        \begin{tabular}{c|c|cccc|cccc}
            \toprule
            \multirow{2}{*}{Model} & \multirow{2}{*}{\makecell{Backbone FLOPs\\{$\downarrow$} (\%)}} & \multicolumn{4}{c|}{Faster-RCNN on VOC} & \multicolumn{4}{c}{Mask-RCNN on COCO}                                                                                                                                                  \\
            \cmidrule{3-10}

                                   &                                                                                       & $mAP$                                       & $AP_{50}$                                 & GFLOPs                         & FPS (img/s) & $mAP^{bbox}$           & $mAP^{segm}$           & GFLOPs                          & FPS (img/s) \\
            \midrule
            Base                   & \multirow{2}{*}{42.32}                                                                & 75.34                                       & 75.30                                     & 109                            & 40        & 36.5                   & 33.0                   & 251                             & 26        \\
            Base + KDFS            &                                                                                       & 76.07 (\textbf{0.73}$\uparrow$)             & 76.10 (\textbf{0.80}$\uparrow$)           & 72 (\textbf{34}\%$\downarrow$) &      44       & 36.2 (0.3$\downarrow$) & 32.8 (0.2$\downarrow$) & 180 (\textbf{28}\%$\downarrow$) &    29         \\
            \bottomrule
        \end{tabular}
    }
    \label{table12}
\end{table*}



\subsubsection{Effect of differential filter sampler}
We further analyze the effect under the same pruned network architectures. In Table~\ref{table11}, Random and Scratch have the same network architectures as ours, so as to achieve the same compression rate. Different from Random and Scratch, our KDFS is able to automatically adjust the important filters during training, while Random and Scratch are kept fixed. First, Scratch achieves higher accuracy than the random filter selection strategy, which is due to the initial filter guidance of our method. In addition, we observe the following findings: (1) The effectiveness of differential filter sampler. Compared to the random method, the only difference is that our KDFS uses the differential filter sampler to automatically adjust the important filters and the sampler significantly improves the accuracy by at least $0.22\%$. (2) with an increase in the compression rate, the accuracy gap becomes smaller compared to both random and scratch methods. This finding also explains the potential problems in Fig.~\ref{fig4} at the large compression rate.

\subsection{Downstream Task Transferring}
\label{other tasks}
To evaluate the strong generalization ability of the pruned models obtained by KDFS, we apply our pruned ResNet-50 for the tasks of object detection and instance segmentation. We train the base and pruned models both on the PASCAL VOC 2012+2007 and MS COCO datasets using Faster-RCNN-FPN~\cite{ren2015faster} and Mask-RCNN-FPN~\cite{he2017mask} to evaluate the performance for object detection and instance segmentation, respectively. In our experiments, we implement the training and inference via the MMDetection package~\footnote{https://github.com/open-mmlab/mmdetection}. The learning strategy and hyper-parameters are the default settings of the original papers. For object detection, we evaluate the detection performance on the PSACAL VOC07 test set.

The results are summarized in Table~\ref{table12}.
For object detection, our pruned model achieves $0.73\%$ mAP and $0.8\%$ $AP_{50}$ improvement while reducing $34\%$ GFLOPs and increasing $4$ FPS, compared to the original Faster-RCNN. For instance segmentation, with a minor segmentation mAP dropping, our method reduces $28\%$ GFLOPs and increases $3$ FPS, compared to the original Mask-RCNN.
It indicates that our KDFS can be effectively transferred to downstream tasks, such as object detection and instance segmentation.

\section{Conclusion}
\label{con}

In this paper, we propose a knowledge-driven differential filter sampler~(KDFS) to globally prune the redundant filters learned from the knowledge of the pre-trained model in a differential and non-alternative optimization. Specifically, we design a differential sampler to build a binary mask vector for each layer, which automatically determines the important filters.
To effectively implement optimization, we exploit the intermediate knowledge from the pre-trained model and propose masked filter modeling to construct PCA-like knowledge, which better guides the learning of pruned networks via the Gumbel-Softmax Straight-Through Gradient Estimator. Extensive experiments on different network architectures and datasets evaluate the effectiveness of KDFS for image classification, as well as strong generalization ability for downstream detection and instance segmentation.

\section*{Acknowledgements}
This work is supported by the National Natural Science Foundation of China (NO. 62102151, NO. 62171391), Shanghai Sailing Program (21YF1411200), Beijing Natural Science Foundation (L223024), Shanghai Science and Technology Commission (22511104600), the Open Research Fund of Key Laboratory of Advanced Theory and Application in Statistics and Data Science, Ministry of Education (KLATASDS2305) and the Fundamental Research Funds for the Central Universities.


\bibliographystyle{IEEEtran}

\begin{thebibliography}{100}
\providecommand{\url}[1]{#1}
\csname url@samestyle\endcsname
\providecommand{\newblock}{\relax}
\providecommand{\bibinfo}[2]{#2}
\providecommand{\BIBentrySTDinterwordspacing}{\spaceskip=0pt\relax}
\providecommand{\BIBentryALTinterwordstretchfactor}{4}
\providecommand{\BIBentryALTinterwordspacing}{\spaceskip=\fontdimen2\font plus
\BIBentryALTinterwordstretchfactor\fontdimen3\font minus
  \fontdimen4\font\relax}
\providecommand{\BIBforeignlanguage}[2]{{%
\expandafter\ifx\csname l@#1\endcsname\relax
\typeout{** WARNING: IEEEtran.bst: No hyphenation pattern has been}%
\typeout{** loaded for the language `#1'. Using the pattern for}%
\typeout{** the default language instead.}%
\else
\language=\csname l@#1\endcsname
\fi
#2}}
\providecommand{\BIBdecl}{\relax}
\BIBdecl

\bibitem{krizhevsky2012imagenet}
A.~Krizhevsky, I.~Sutskever, and G.~E. Hinton, ``Imagenet classification with
  deep convolutional neural networks,'' in \emph{NeurIPS}, 2012, pp.
  1097--1105.

\bibitem{simonyan2015very}
K.~Simonyan and A.~Zisserman, ``Very deep convolutional networks for
  large-scale image recognition,'' in \emph{ICLR}, 2015.

\bibitem{szegedy2015going}
C.~Szegedy, W.~Liu, Y.~Jia, P.~Sermanet, S.~Reed, D.~Anguelov, D.~Erhan,
  V.~Vanhoucke, and A.~Rabinovich, ``Going deeper with convolutions,'' in
  \emph{CVPR}, 2015, pp. 1--9.

\bibitem{he2016deep}
K.~He, X.~Zhang, S.~Ren, and J.~Sun, ``Deep residual learning for image
  recognition,'' in \emph{CVPR}, 2016, pp. 770--778.

\bibitem{huang2017densely}
G.~Huang, Z.~Liu, L.~van~der Maaten, and K.~Q. Weinberger, ``Densely connected
  convolutional networks,'' in \emph{CVPR}, 2017, pp. 4700--4708.

\bibitem{girshick2014rich}
R.~Girshick, J.~Donahue, T.~Darrell, and J.~Malik, ``Rich feature hierarchies
  for accurate object detection and semantic segmentation,'' in \emph{CVPR},
  2014, pp. 580--587.

\bibitem{girshick2015fast}
R.~Girshick, ``Fast r-cnn,'' in \emph{ICCV}, 2015, pp. 1440--1448.

\bibitem{ren2015faster}
S.~Ren, K.~He, R.~Girshick, and J.~Sun, ``Faster r-cnn: Towards real-time
  object detection with region proposal networks,'' in \emph{NeurIPS}, 2015,
  pp. 91--99.

\bibitem{he2017mask}
K.~He, G.~Gkioxari, P.~Doll{\'a}r, and R.~Girshick, ``Mask r-cnn,'' in
  \emph{ICCV}, 2017, pp. 2961--2969.

\bibitem{long2015fully}
J.~Long, E.~Shelhamer, and T.~Darrell, ``Fully convolutional networks for
  semantic segmentation,'' in \emph{CVPR}, 2015, pp. 3431--3440.

\bibitem{chen2017deeplab}
L.-C. Chen, G.~Papandreou, I.~Kokkinos, K.~Murphy, and A.~L. Yuille, ``Deeplab:
  Semantic image segmentation with deep convolutional nets, atrous convolution,
  and fully connected crfs,'' \emph{TPAMI}, vol.~40, no.~4, pp. 834--848, 2017.

\bibitem{chen2017rethinking}
L.-C. Chen, G.~Papandreou, F.~Schroff, and H.~Adam, ``Rethinking atrous
  convolution for semantic image segmentation,'' \emph{arXiv preprint
  arXiv:1706.05587}, 2017.

\bibitem{lin2018holistic}
S.~Lin, R.~Ji, C.~Chen, D.~Tao, and J.~Luo, ``Holistic cnn compression via
  low-rank decomposition with knowledge transfer,'' \emph{TPAMI}, 2018.

\bibitem{li2021towards}
Y.~Li, S.~Lin, J.~Liu, Q.~Ye, M.~Wang, F.~Chao, F.~Yang, J.~Ma, Q.~Tian, and
  R.~Ji, ``Towards compact cnns via collaborative compression,'' in
  \emph{CVPR}, 2021, pp. 6438--6447.

\bibitem{zhang2015accelerating}
X.~Zhang, J.~Zou, K.~He, and J.~Sun, ``Accelerating very deep convolutional
  networks for classification and detection,'' \emph{TPAMI}, vol.~38, no.~10,
  pp. 1943--1955, 2015.

\bibitem{denton2014exploiting}
E.~L. Denton, W.~Zaremba, J.~Bruna, Y.~LeCun, and R.~Fergus, ``Exploiting
  linear structure within convolutional networks for efficient evaluation,'' in
  \emph{NeurIPS}, 2014, pp. 1269--1277.

\bibitem{hinton2015distilling}
G.~Hinton, O.~Vinyals, and J.~Dean, ``Distilling the knowledge in a neural
  network,'' \emph{arXiv preprint arXiv:1503.02531}, 2015.

\bibitem{romero2014fitnets}
A.~Romero, N.~Ballas, S.~E. Kahou, A.~Chassang, C.~Gatta, and Y.~Bengio,
  ``Fitnets: Hints for thin deep nets,'' \emph{arXiv preprint arXiv:1412.6550},
  2014.

\bibitem{rastegari2016xnor}
M.~Rastegari, V.~Ordonez, J.~Redmon, and A.~Farhadi, ``Xnor-net: Imagenet
  classification using binary convolutional neural networks,'' in \emph{ECCV},
  2016.

\bibitem{jacob2018quantization}
B.~Jacob, S.~Kligys, B.~Chen, M.~Zhu, M.~Tang, A.~Howard, H.~Adam, and
  D.~Kalenichenko, ``Quantization and training of neural networks for efficient
  integer-arithmetic-only inference,'' in \emph{CVPR}, 2018, pp. 2704--2713.

\bibitem{zhao2022towards}
J.~Zhao, S.~Xu, B.~Zhang, J.~Gu, D.~Doermann, and G.~Guo, ``Towards compact
  1-bit cnns via bayesian learning,'' \emph{IJCV}, pp. 1--25, 2022.

\bibitem{han2015learning}
S.~Han, J.~Pool, J.~Tran, and W.~Dally, ``Learning both weights and connections
  for efficient neural network,'' in \emph{NeurIPS}, 2015, pp. 1135--1143.

\bibitem{han2016deep}
S.~Han, H.~Mao, and W.~J. Dally, ``Deep compression: Compressing deep neural
  network with pruning, trained quantization and huffman coding,'' in
  \emph{ICLR}, 2016.

\bibitem{frankle2018lottery}
J.~Frankle and M.~Carbin, ``The lottery ticket hypothesis: Finding sparse,
  trainable neural networks,'' \emph{arXiv preprint arXiv:1803.03635}, 2018.

\bibitem{lin2019towards}
S.~Lin, R.~Ji, C.~Yan, B.~Zhang, L.~Cao, Q.~Ye, F.~Huang, and D.~Doermann,
  ``Towards optimal structured cnn pruning via generative adversarial
  learning,'' in \emph{CVPR}, 2019, pp. 2790--2799.

\bibitem{li2017pruning}
H.~Li, A.~Kadav, I.~Durdanovic, H.~Samet, and H.~P. Graf, ``Pruning filters for
  efficient convnets,'' in \emph{ICLR}, 2017.

\bibitem{liu2017learning}
Z.~Liu, J.~Li, Z.~Shen, G.~Huang, S.~Yan, and C.~Zhang, ``Learning efficient
  convolutional networks through network slimming,'' in \emph{ICCV}, 2017, pp.
  2755--2763.

\bibitem{lin2020hrank}
M.~Lin, R.~Ji, Y.~Wang, Y.~Zhang, B.~Zhang, Y.~Tian, and L.~Shao, ``Hrank:
  Filter pruning using high-rank feature map,'' in \emph{CVPR}, 2020, pp.
  1529--1538.

\bibitem{he2019filter}
Y.~He, P.~Liu, Z.~Wang, Z.~Hu, and Y.~Yang, ``Filter pruning via geometric
  median for deep convolutional neural networks acceleration,'' in \emph{CVPR},
  2019, pp. 4340--4349.

\bibitem{park2017faster}
J.~Park, S.~Li, W.~Wen, P.~T.~P. Tang, H.~Li, Y.~Chen, and P.~Dubey, ``Faster
  cnns with direct sparse convolutions and guided pruning,'' in \emph{ICLR},
  2017.

\bibitem{han2016eie}
S.~Han, X.~Liu, H.~Mao, J.~Pu, A.~Pedram, M.~A. Horowitz, and W.~J. Dally,
  ``Eie: efficient inference engine on compressed deep neural network,'' in
  \emph{ISCA}, 2016, pp. 243--254.

\bibitem{he2017channel}
Y.~He, X.~Zhang, and J.~Sun, ``Channel pruning for accelerating very deep
  neural networks,'' in \emph{ICCV}, 2017, pp. 1389--1397.

\bibitem{luo2017thinet}
J.-H. Luo, J.~Wu, and W.~Lin, ``Thinet: A filter level pruning method for deep
  neural network compression,'' in \emph{ICCV}, 2017, pp. 5058--5066.

\bibitem{hu2016network}
H.~Hu, R.~Peng, Y.-W. Tai, and C.-K. Tang, ``Network trimming: A data-driven
  neuron pruning approach towards efficient deep architectures,'' \emph{arXiv
  preprint arXiv:1607.03250}, 2016.

\bibitem{lin2019toward}
S.~Lin, R.~Ji, Y.~Li, C.~Deng, and X.~Li, ``Toward compact convnets via
  structure-sparsity regularized filter pruning,'' \emph{TNNLS}, vol.~31,
  no.~2, pp. 574--588, 2019.

\bibitem{zhou2016less}
H.~Zhou, J.~M. Alvarez, and F.~Porikli, ``Less is more: Towards compact cnns,''
  in \emph{ECCV}, 2016, pp. 662--677.

\bibitem{wen2016learning}
W.~Wen, C.~Wu, Y.~Wang, Y.~Chen, and H.~Li, ``Learning structured sparsity in
  deep neural networks,'' in \emph{NeurIPS}, 2016.

\bibitem{ye2018rethinking}
J.~Ye, X.~Lu, Z.~Lin, and J.~Z. Wang, ``Rethinking the
  smaller-norm-less-informative assumption in channel pruning of convolution
  layers,'' in \emph{ICLR}, 2018.

\bibitem{you2019gate}
Z.~You, K.~Yan, J.~Ye, M.~Ma, and P.~Wang, ``Gate decorator: Global filter
  pruning method for accelerating deep convolutional neural networks,'' in
  \emph{NeurIPS}, vol.~32, 2019.

\bibitem{huang2018data}
Z.~Huang and N.~Wang, ``Data-driven sparse structure selection for deep neural
  networks,'' in \emph{ECCV}, 2018.

\bibitem{yu2018nisp}
R.~Yu, A.~Li, C.-F. Chen, J.-H. Lai, V.~I. Morariu, X.~Han, M.~Gao, C.-Y. Lin,
  and L.~S. Davis, ``Nisp: Pruning networks using neuron importance score
  propagation,'' in \emph{CVPR}, 2018, pp. 9194--9203.

\bibitem{gao2022disentangled}
S.~Gao, F.~Huang, Y.~Zhang, and H.~Huang, ``Disentangled differentiable network
  pruning,'' in \emph{ECCV}.\hskip 1em plus 0.5em minus 0.4em\relax Springer,
  2022, pp. 328--345.

\bibitem{lin2018accelerating}
S.~Lin, R.~Ji, Y.~Li, Y.~Wu, F.~Huang, and B.~Zhang, ``Accelerating
  convolutional networks via global \& dynamic filter pruning.'' in
  \emph{IJCAI}, 2018.

\bibitem{gao2020discrete}
S.~Gao, F.~Huang, J.~Pei, and H.~Huang, ``Discrete model compression with
  resource constraint for deep neural networks,'' in \emph{CVPR}, 2020, pp.
  1899--1908.

\bibitem{guo2021gdp}
Y.~Guo, H.~Yuan, J.~Tan, Z.~Wang, S.~Yang, and J.~Liu, ``Gdp: Stabilized neural
  network pruning via gates with differentiable polarization,'' in \emph{ICCV},
  2021, pp. 5239--5250.

\bibitem{he2022masked}
K.~He, X.~Chen, S.~Xie, Y.~Li, P.~Doll{\'a}r, and R.~Girshick, ``Masked
  autoencoders are scalable vision learners,'' in \emph{CVPR}, 2022, pp.
  16\,000--16\,009.

\bibitem{hassibi1993second}
B.~Hassibi and D.~G. Stork, ``Second order derivatives for network pruning:
  Optimal brain surgeon,'' in \emph{NeurIPS}, 1993.

\bibitem{lecun1989optimal}
Y.~LeCun, J.~S. Denker, S.~A. Solla, R.~E. Howard, and L.~D. Jackel, ``Optimal
  brain damage.'' in \emph{NeurIPS}, vol.~2, 1989, pp. 598--605.

\bibitem{guo2016dynamic}
Y.~Guo, A.~Yao, and Y.~Chen, ``Dynamic network surgery for efficient dnns,'' in
  \emph{NeurIPS}, 2016, pp. 1379--1387.

\bibitem{molchanov2017pruning}
P.~Molchanov, S.~Tyree, T.~Karras, T.~Aila, and J.~Kautz, ``Pruning
  convolutional neural networks for resource efficient inference,'' in
  \emph{ICLR}, 2017.

\bibitem{molchanov2019importance}
P.~Molchanov, A.~Mallya, S.~Tyree, I.~Frosio, and J.~Kautz, ``Importance
  estimation for neural network pruning,'' in \emph{CVPR}, 2019, pp.
  11\,264--11\,272.

\bibitem{dong2017learning}
X.~Dong, S.~Chen, and S.~Pan, ``Learning to prune deep neural networks via
  layer-wise optimal brain surgeon,'' in \emph{NeurIPS}, 2017.

\bibitem{peng2019collaborative}
H.~Peng, J.~Wu, S.~Chen, and J.~Huang, ``Collaborative channel pruning for deep
  networks,'' in \emph{ICML}, 2019, pp. 5113--5122.

\bibitem{zheng2022savit}
C.~Zheng, K.~Zhang, Z.~Yang, W.~Tan, J.~Xiao, Y.~Ren, S.~Pu \emph{et~al.},
  ``Savit: Structure-aware vision transformer pruning via collaborative
  optimization,'' in \emph{NeurIPS}, 2022.

\bibitem{bengio2013estimating}
Y.~Bengio, N.~L{\'e}onard, and A.~Courville, ``Estimating or propagating
  gradients through stochastic neurons for conditional computation,''
  \emph{arXiv preprint arXiv:1308.3432}, 2013.

\bibitem{hu2018squeeze}
J.~Hu, L.~Shen, and G.~Sun, ``Squeeze-and-excitation networks,'' in
  \emph{CVPR}, 2018, pp. 7132--7141.

\bibitem{veit2018convolutional}
A.~Veit and S.~Belongie, ``Convolutional networks with adaptive inference
  graphs,'' in \emph{ECCV}, 2018, pp. 3--18.

\bibitem{chen2019you}
Z.~Chen, Y.~Li, S.~Bengio, and S.~Si, ``You look twice: Gaternet for dynamic
  filter selection in cnns,'' in \emph{CVPR}, 2019, pp. 9172--9180.

\bibitem{gao2018dynamic}
X.~Gao, Y.~Zhao, {\L}.~Dudziak, R.~Mullins, and C.-z. Xu, ``Dynamic channel
  pruning: Feature boosting and suppression,'' \emph{arXiv preprint
  arXiv:1810.05331}, 2018.

\bibitem{liu2018rethinking}
Z.~Liu, M.~Sun, T.~Zhou, G.~Huang, and T.~Darrell, ``Rethinking the value of
  network pruning,'' \emph{arXiv preprint arXiv:1810.05270}, 2018.

\bibitem{he2018amc}
Y.~He, J.~Lin, Z.~Liu, H.~Wang, L.-J. Li, and S.~Han, ``Amc: Automl for model
  compression and acceleration on mobile devices,'' in \emph{ECCV}, 2018, pp.
  784--800.

\bibitem{lin2021channel}
M.~Lin, R.~Ji, Y.~Zhang, B.~Zhang, Y.~Wu, and Y.~Tian, ``Channel pruning via
  automatic structure search,'' in \emph{IJCAI}, 2021, pp. 673--679.

\bibitem{lin2017runtime}
J.~Lin, Y.~Rao, J.~Lu, and J.~Zhou, ``Runtime neural pruning,'' in
  \emph{NeurIPS}, 2017.

\bibitem{liu2019knowledge}
Y.~Liu, J.~Cao, B.~Li, C.~Yuan, W.~Hu, Y.~Li, and Y.~Duan, ``Knowledge
  distillation via instance relationship graph,'' in \emph{CVPR}, 2019, pp.
  7096--7104.

\bibitem{ji2021show}
M.~Ji, B.~Heo, and S.~Park, ``Show, attend and distill: Knowledge distillation
  via attention-based feature matching,'' in \emph{AAAI}, vol.~35, no.~9, 2021,
  pp. 7945--7952.

\bibitem{zagoruyko2016paying}
S.~Zagoruyko and N.~Komodakis, ``Paying more attention to attention: Improving
  the performance of convolutional neural networks via attention transfer,''
  \emph{arXiv preprint arXiv:1612.03928}, 2016.

\bibitem{micaelli2019zero}
P.~Micaelli and A.~J. Storkey, ``Zero-shot knowledge transfer via adversarial
  belief matching,'' in \emph{NeurIPS}, 2019.

\bibitem{zhuang2018discrimination}
Z.~Zhuang, M.~Tan, B.~Zhuang, J.~Liu, Y.~Guo, Q.~Wu, J.~Huang, and J.~Zhu,
  ``Discrimination-aware channel pruning for deep neural networks,'' in
  \emph{NeurIPS}, 2018.

\bibitem{he2015delving}
K.~He, X.~Zhang, S.~Ren, and J.~Sun, ``Delving deep into rectifiers: Surpassing
  human-level performance on imagenet classification,'' in \emph{ICCV}, 2015,
  pp. 1026--1034.

\bibitem{jang2016categorical}
E.~Jang, S.~Gu, and B.~Poole, ``Categorical reparameterization with
  gumbel-softmax,'' \emph{arXiv preprint arXiv:1611.01144}, 2016.

\bibitem{yim2017gift}
J.~Yim, D.~Joo, J.~Bae, and J.~Kim, ``A gift from knowledge distillation: Fast
  optimization, network minimization and transfer learning,'' in \emph{CVPR},
  2017, pp. 4133--4141.

\bibitem{kingma2014adam}
D.~P. Kingma and J.~Ba, ``Adam: A method for stochastic optimization,''
  \emph{arXiv preprint arXiv:1412.6980}, 2014.

\bibitem{chin2020towards}
T.-W. Chin, R.~Ding, C.~Zhang, and D.~Marculescu, ``Towards efficient model
  compression via learned global ranking,'' in \emph{CVPR}, 2020, pp.
  1518--1528.

\bibitem{lin2021filter}
M.~Lin, L.~Cao, S.~Li, Q.~Ye, Y.~Tian, J.~Liu, Q.~Tian, and R.~Ji, ``Filter
  sketch for network pruning,'' \emph{TNNLS}, vol.~33, no.~12, pp. 7091--7100,
  2021.

\bibitem{kim2019efficient}
H.~Kim, M.~U.~K. Khan, and C.-M. Kyung, ``Efficient neural network
  compression,'' in \emph{CVPR}, 2019, pp. 12\,569--12\,577.

\bibitem{zhuang2020neuron}
T.~Zhuang, Z.~Zhang, Y.~Huang, X.~Zeng, K.~Shuang, and X.~Li, ``Neuron-level
  structured pruning using polarization regularizer,'' in \emph{NeurIPS}, 2020.

\bibitem{ning2020dsa}
X.~Ning, T.~Zhao, W.~Li, P.~Lei, Y.~Wang, and H.~Yang, ``Dsa: More efficient
  budgeted pruning via differentiable sparsity allocation,'' in
  \emph{ECCV}.\hskip 1em plus 0.5em minus 0.4em\relax Springer, 2020, pp.
  592--607.

\bibitem{li2020group}
Y.~Li, S.~Gu, C.~Mayer, L.~V. Gool, and R.~Timofte, ``Group sparsity: The hinge
  between filter pruning and decomposition for network compression,'' in
  \emph{CVPR}, 2020, pp. 8018--8027.

\bibitem{gao2021network}
S.~Gao, F.~Huang, W.~Cai, and H.~Huang, ``Network pruning via performance
  maximization,'' in \emph{CVPR}, 2021, pp. 9270--9280.

\bibitem{he2018soft}
Y.~He, G.~Kang, X.~Dong, Y.~Fu, and Y.~Yang, ``Soft filter pruning for
  accelerating deep convolutional neural networks,'' \emph{arXiv preprint
  arXiv:1808.06866}, 2018.

\bibitem{dong2019network}
X.~Dong and Y.~Yang, ``Network pruning via transformable architecture search,''
  in \emph{NeurIPS}, 2019.

\bibitem{kang2020operation}
M.~Kang and B.~Han, ``Operation-aware soft channel pruning using differentiable
  masks,'' in \emph{ICML}.\hskip 1em plus 0.5em minus 0.4em\relax PMLR, 2020,
  pp. 5122--5131.

\bibitem{tang2020scop}
Y.~Tang, Y.~Wang, Y.~Xu, D.~Tao, C.~Xu, C.~Xu, and C.~Xu, ``Scop: Scientific
  control for reliable neural network pruning,'' in \emph{NeurIPS}, 2020.

\bibitem{le2021network}
D.~H. Le and B.-S. Hua, ``Network pruning that matters: A case study on
  retraining variants,'' \emph{arXiv preprint arXiv:2105.03193}, 2021.

\bibitem{zhang2021filter}
X.~Zhang, W.~Xie, Y.~Li, J.~Lei, and Q.~Du, ``Filter pruning via learned
  representation median in the frequency domain,'' \emph{TCYB}, 2021.

\bibitem{ruan2021dpfps}
X.~Ruan, Y.~Liu, B.~Li, C.~Yuan, and W.~Hu, ``Dpfps: dynamic and progressive
  filter pruning for compressing convolutional neural networks from scratch,''
  in \emph{AAAI}, vol.~35, no.~3, 2021, pp. 2495--2503.

\bibitem{jiang2023pruning}
N.-F. Jiang, X.~Zhao, C.-Y. Zhao, Y.-Q. An, M.~Tang, and J.-Q. Wang,
  ``Pruning-aware sparse regularization for network pruning,'' \emph{Machine
  Intelligence Research}, vol.~20, no.~1, pp. 109--120, 2023.

\bibitem{he2020learning}
Y.~He, Y.~Ding, P.~Liu, L.~Zhu, H.~Zhang, and Y.~Yang, ``Learning filter
  pruning criteria for deep convolutional neural networks acceleration,'' in
  \emph{CVPR}, 2020, pp. 2009--2018.

\bibitem{krizhevsky2009learning}
A.~Krizhevsky and G.~Hinton, ``Learning multiple layers of features from tiny
  images,'' Citeseer, Tech. Rep., 2009.

\bibitem{russakovsky2015imagenet}
O.~Russakovsky, J.~Deng, H.~Su, J.~Krause, S.~Satheesh, S.~Ma, Z.~Huang,
  A.~Karpathy, A.~Khosla, M.~Bernstein \emph{et~al.}, ``Imagenet large scale
  visual recognition challenge,'' \emph{IJCV}, vol. 115, no.~3, pp. 211--252,
  2015.

\bibitem{sandler2018mobilenetv2}
M.~Sandler, A.~Howard, M.~Zhu, A.~Zhmoginov, and L.-C. Chen, ``Mobilenetv2:
  Inverted residuals and linear bottlenecks,'' in \emph{CVPR}, 2018, pp.
  4510--4520.

\bibitem{pascal-voc-2012}
M.~Everingham, L.~Van~Gool, C.~K.~I. Williams, J.~Winn, and A.~Zisserman, ``The
  {PASCAL} {V}isual {O}bject {C}lasses {C}hallenge 2012 {(VOC2012)}
  {R}esults,''
  http://www.pascal-network.org/challenges/VOC/voc2012/workshop/index.html.

\bibitem{lin2014microsoft}
T.-Y. Lin, M.~Maire, S.~Belongie, J.~Hays, P.~Perona, D.~Ramanan,
  P.~Doll{\'a}r, and C.~L. Zitnick, ``Microsoft coco: Common objects in
  context,'' in \emph{ECCV}.\hskip 1em plus 0.5em minus 0.4em\relax Springer,
  2014, pp. 740--755.

\bibitem{paszke2019pytorch}
A.~Paszke, S.~Gross, F.~Massa, A.~Lerer, J.~Bradbury, G.~Chanan, T.~Killeen,
  Z.~Lin, N.~Gimelshein, L.~Antiga \emph{et~al.}, ``Pytorch: An imperative
  style, high-performance deep learning library,'' in \emph{NeurIPS}, 2019.

\bibitem{zhao2019variational}
C.~Zhao, B.~Ni, J.~Zhang, Q.~Zhao, W.~Zhang, and Q.~Tian, ``Variational
  convolutional neural network pruning,'' in \emph{CVPR}, 2019, pp. 2780--2789.

\bibitem{lin2022ezcrop}
R.~Lin, J.~Ran, D.~Wang, K.~H. Chiu, and N.~Wong, ``Ezcrop: Energy-zoned
  channels for robust output pruning,'' in \emph{WACV}, 2022, pp. 19--28.

\bibitem{ding2020prune}
G.~Ding, S.~Zhang, Z.~Jia, J.~Zhong, and J.~Han, ``Where to prune: Using lstm
  to guide data-dependent soft pruning,'' \emph{TIP}, vol.~30, pp. 293--304,
  2020.

\bibitem{wang2021convolutional}
Z.~Wang, C.~Li, and X.~Wang, ``Convolutional neural network pruning with
  structural redundancy reduction,'' in \emph{CVPR}, 2021, pp.
  14\,913--14\,922.

\bibitem{guo2020dmcp}
S.~Guo, Y.~Wang, Q.~Li, and J.~Yan, ``Dmcp: Differentiable markov channel
  pruning for neural networks,'' in \emph{CVPR}, 2020, pp. 1539--1547.

\bibitem{joo2021linearly}
D.~Joo, E.~Yi, S.~Baek, and J.~Kim, ``Linearly replaceable filters for deep
  network channel pruning,'' in \emph{AAAI}, vol.~35, no.~9, 2021, pp.
  8021--8029.

\bibitem{qian2023hierarchical}
Y.~Qian, Z.~He, Y.~Wang, B.~Wang, X.~Ling, Z.~Gu, H.~Wang, S.~Zeng, and
  W.~Swaileh, ``Hierarchical threshold pruning based on uniform response
  criterion,'' \emph{TNNLS}, 2023.

\bibitem{liu2019metapruning}
Z.~Liu, H.~Mu, X.~Zhang, Z.~Guo, X.~Yang, K.-T. Cheng, and J.~Sun,
  ``Metapruning: Meta learning for automatic neural network channel pruning,''
  in \emph{ICCV}, 2019, pp. 3296--3305.

\bibitem{ye2020good}
M.~Ye, C.~Gong, L.~Nie, D.~Zhou, A.~Klivans, and Q.~Liu, ``Good subnetworks
  provably exist: Pruning via greedy forward selection,'' in \emph{ICML}.\hskip
  1em plus 0.5em minus 0.4em\relax PMLR, 2020, pp. 10\,820--10\,830.

\bibitem{liu2021joint}
Z.~Liu, X.~Zhang, Z.~Shen, Y.~Wei, K.-T. Cheng, and J.~Sun, ``Joint
  multi-dimension pruning via numerical gradient update,'' \emph{TIP}, vol.~30,
  pp. 8034--8045, 2021.

\bibitem{guo2020model}
J.~Guo, W.~Zhang, W.~Ouyang, and D.~Xu, ``Model compression using progressive
  channel pruning,'' \emph{TCSVT}, vol.~31, no.~3, pp. 1114--1124, 2020.

\bibitem{zheng2022model}
Y.-J. Zheng, S.-B. Chen, C.~H. Ding, and B.~Luo, ``Model compression based on
  differentiable network channel pruning,'' \emph{TNNLS}, 2022.

\end{thebibliography}
\end{document}